\documentclass[journal]{IEEEtran}

\usepackage{cite}
\usepackage{amsmath,amsfonts,amssymb}

\usepackage{graphicx}
\usepackage{textcomp}
\usepackage{xcolor}
\usepackage{booktabs}      
\usepackage{array}          
\usepackage{float}          
\usepackage{enumitem}       
\usepackage{rotating}       

\usepackage{hyperref}       
\usepackage{url}

\usepackage[utf8]{inputenc}

\hypersetup{
    pdfencoding=auto,
    bookmarks=false,
    colorlinks=true,
    linkcolor=blue,
    citecolor=blue,
    urlcolor=blue,
    pdftitle={MTEB-BR: A Text Embedding Benchmark for Brazilian Portuguese},
    pdfauthor={Tardelli Ronan Coelho Stekel},
}

\graphicspath{{figures/}}

\begin{document}

\title{MTEB-BR: A Text Embedding Benchmark for Brazilian Portuguese}

\author{%
    Tardelli Ronan Coelho Stekel\\[2pt]
    Federal Institute of São Paulo (IFSP)\\
    São Paulo, Brazil\\
    stekel@ifsp.edu.br
}

\maketitle
\thispagestyle{empty}

\begin{abstract}
Text embeddings for Portuguese have no dedicated benchmark: evaluation rests on translated corpora such as English MS MARCO or on thin multilingual coverage, with native tasks scattered and unconsolidated. We introduce MTEB-BR, a benchmark of 22 native Brazilian-Portuguese tasks across seven categories (classification, multilabel classification, pair classification, semantic textual similarity, clustering, retrieval, and reranking), admitting only data created or found in Portuguese and excluding translations by construction. We evaluate 93 models spanning 23M to 27B parameters: 73 open-weight and 20 closed commercial APIs. Alongside the leaderboard we report a statistical layer for every headline comparison: per-task bootstrap confidence intervals, paired-bootstrap significance, a task- and instance-level discrimination analysis (how sharply each task separates models) adapted from Item Response Theory, and a cross-leaderboard correlation. Three findings stand out. The benchmark cleanly separates about a dozen tiers of models, though the top six are statistically too close to order. An openly licensed, self-hostable model reaches that leading tier, so strong Portuguese embedding quality does not require a commercial API. And a model's rank on the global multilingual leaderboard predicts its Portuguese rank only moderately (Spearman $\rho = 0.75$ over 55 shared models; one model ranks 3rd there and 49th here), so a native benchmark measures something the multilingual boards do not. We release every task, our code, and a public leaderboard, so practitioners can choose Portuguese embedding models on native evidence.

\end{abstract}

\section{Introduction}\label{sec:intro}

Brazilian Portuguese is spoken natively by over 200 million people, yet a practitioner who needs to deploy a sentence-embedding model for it, whether for semantic search, classification, or retrieval-augmented generation, has no comprehensive native benchmark to guide that choice. The dominant evaluation frameworks, MTEB \cite{muennighoff-etal-2023-mteb} and its multilingual successor MMTEB \cite{enevoldsen-etal-2025-mmteb}, do include Portuguese tasks, but they are a small fraction of MMTEB's 500-task suite, and the most heavily reported Portuguese retrieval task is mMARCO-PT \cite{bonifacio-etal-2021-mmarco}, a machine translation of English MS MARCO. Translation introduces systematic artifacts (translation noise, domain drift, idiomatic flattening) that can mask real differences between models on native text \cite{pham-etal-2025-vn-mteb}.

Language-specific MTEB extensions have closed this gap for Chinese \cite{xiao-etal-2024-cmteb}, Scandinavian \cite{enevoldsen-etal-2024-scandeval}, French \cite{ciancone-etal-2024-fr-mteb}, Polish \cite{poswiata-etal-2024-polish-mteb}, Russian \cite{snegirev-etal-2024-rumteb}, Persian \cite{zinvandi-etal-2025-famteb}, Dutch \cite{banar-etal-2025-mteb-nl}, German \cite{wehrli-etal-2024-german-clust}, Vietnamese \cite{pham-etal-2025-vn-mteb}, and Japanese \cite{tsukagoshi-etal-2024-ruri-jmteb}, but no comprehensive native one existed for Portuguese. We close that gap, and add a layer of statistical rigor that complements the established MTEB methodology. We make four contributions:

\begin{enumerate}
\item \textbf{A native, non-translated task suite.} We curate 22 tasks across seven MTEB categories from existing Brazilian-Portuguese resources spanning legal, medical, tax, scientific, encyclopedic, and social-media domains, excluding machine-translated benchmarks (notably mMARCO-PT and mkqa-PT) by construction.
\item \textbf{A large and diverse model panel.} We evaluate 93 models: 73 open-weight (23M--27B parameters) and 20 closed commercial APIs, covering the proprietary embedding services alongside the open ecosystem.
\item \textbf{A statistical-rigor layer.} For every headline comparison we report per-task bootstrap confidence intervals, paired-bootstrap $p$-values, Item Response Theory discrimination at the task and instance levels, and a Borda-count robustness check. This layer shows the benchmark cleanly orders $78.7\%$ of all model pairs into about a dozen distinguishable tiers and places the leader above 87 of 92 models, while the top six converge into one unresolved frontier tier (\S\ref{sec:results}).
\item \textbf{A cross-leaderboard agreement analysis.} We correlate MTEB-BR against the live HuggingFace MTEB multilingual leaderboard ($\rho = 0.75$, over the $n = 55$ models common to both) and find it only a moderate predictor of Portuguese rank: one model ranks 3rd of those 55 on that board but 49th of them here (\S\ref{sec:cross_leaderboard}).
\end{enumerate}

All code (Apache-2.0, archived at \url{https://doi.org/10.5281/zenodo.21087216}), the full results dataset (CC-BY 4.0, archived at \url{https://doi.org/10.57967/hf/9491}), and an interactive leaderboard\footnote{\url{https://huggingface.co/spaces/mteb-br/leaderboard}} are released openly.

\section{Related Work}\label{sec:related}

Embedding evaluation has converged on a lineage of benchmarks of growing scope: SentEval \cite{conneau-kiela-2018-senteval} for STS and classification probes, BEIR \cite{thakur-etal-2021-beir} for heterogeneous retrieval, MTEB \cite{muennighoff-etal-2023-mteb} for eight task categories over 58 datasets, and MMTEB \cite{enevoldsen-etal-2025-mmteb} for over 500 tasks across 250+ languages. MMTEB introduced the Borda aggregation and the inter-task-correlation task-selection device we build on, though its per-language coverage for lower-resource languages remains limited, Portuguese among them.

These language-specific extensions vary in scope and rigor: as originally published, most reported on the order of 40--50 models, and few included more than one or two closed commercial APIs. Closer to the retrieval-only tradition, MIRACL \cite{zhang-etal-2023-miracl} offers native multilingual retrieval over 18 languages (Portuguese not among them), and BEIR-PL \cite{wojtasik-etal-2024-beir-pl} exemplifies the translate-and-evaluate workflow we avoid. Portuguese also has prior benchmark collections, but they evaluate task-fine-tuned language models rather than embeddings: Napolab \cite{rodrigues-2023-napolab} gathers native NLU datasets (entailment, similarity, hate speech), and PORTULAN ExtraGLUE \cite{osorio-etal-2024-extraglue} machine-translates the GLUE suite into Portuguese. Neither measures the frozen sentence embeddings MTEB-BR compares. Okamura et al.\ \cite{okamura-etal-2026-mtebpt} proposed a Portuguese sentence-encoder benchmark constructed from MMTEB subsets (14 datasets, 17 models, with additional fine-tuned encoders). Our benchmark differs in both construction and scale: its 22 tasks are natively Brazilian-Portuguese by design, with machine-translated datasets excluded, and it evaluates 93 models spanning both open-weight and closed commercial systems, reported with per-comparison confidence intervals, paired-bootstrap significance, and item-response-theory task discrimination.

These benchmarks have prioritized scope and coverage, with statistical reporting a secondary concern. Several have begun to add it: the Scandinavian Embedding Benchmark (SEB) \cite{enevoldsen-etal-2024-scandeval} reports per-model rank confidence intervals, MTEB-French reports Critical Difference diagrams, and MMTEB introduced inter-task correlation as a task-selection device. We build on this direction for Brazilian Portuguese by reporting four complementary statistics together: per-task confidence intervals, paired-bootstrap significance across model pairs, IRT per-task discrimination, and a quantitative cross-leaderboard correlation.

\section{Benchmark Design}\label{sec:benchmark}

The benchmark's single inclusion criterion is native source. Every task draws on a corpus produced for, or naturally occurring in, Brazilian Portuguese, whether human-created, naturally found, or LM-generated and verified.\footnote{These correspond to the values of the \texttt{sample\_creation} field in the MTEB metadata schema.} Machine-translated corpora are excluded by construction: translation noise, domain drift, and idiomatic flattening introduce artifacts that can mask real differences between models. Within that constraint we select for domain diversity, covering settings where an embedding model is realistically deployed: social-media classification, legal and tax retrieval, medical retrieval and clustering, open-domain retrieval, and semantic similarity and inference. The filter admits one LM-generated task, ToxSynPT, synthesized and human-verified directly in Brazilian Portuguese; synthetic generation may nonetheless introduce artifacts of its own.

The 22 headline tasks span seven MTEB categories: 4 classification, 1 multilabel classification, 2 pair classification, 2 STS, 5 clustering, 6 retrieval, and 2 reranking; these 22 formulations draw on 17 distinct corpora, as four sources (ASSIN, Quati, JurisTCU, MedPT) recur under multiple protocols. The retrieval weighting (6 of 22, against 1 of 8 in the original MTEB) reflects both the practical centrality of search and retrieval-augmented generation and the discrimination analysis of \S\ref{sec:irt}, where retrieval is the category that most sharply separates models; it is carried by four retrieval corpora beyond the reused Quati and JurisTCU sources (MedPTRetrieval, FaqBacen, FaQuAD-IR, BR-TaxQA-R). Table~\ref{tab:per_task} gives each task's source dataset, license, evaluation-set size, corpus or label structure, and citation.

\begin{table*}[t]
\centering
\caption{Per-task metadata and evaluation-set statistics for the 22 MTEB-BR tasks. \emph{Eval} = test instances, sentence pairs, texts, or queries, by task type. \emph{Corpus / labels} = classes, clusters, corpus documents, or judged candidates; classification shows the linear-probe train size in parentheses.}
\label{tab:per_task}
\small
\resizebox{\textwidth}{!}{%
\begin{tabular}{llllrll}
\toprule
Task & Category & Source dataset (HF) & License & Eval & Corpus / labels & Citation \\
\midrule
HateBR & Classification & \texttt{franciellevargas/HateBR} & apache-2.0 & 1{,}400 & 2 classes (5{,}600 train) & \cite{vargas-etal-2022-hatebr} \\
FactckBrClassification & Classification & \texttt{mteb-br/factckbr} & apache-2.0 & 259 & 3 classes (1{,}029 train) & \cite{moreno2019factckbr} \\
ToxSynPT & Classification & \texttt{AKCIT/ToxSyn-PT} & cc-by-4.0 & 5{,}208 & 2 classes (48{,}066 train) & \cite{brito-etal-2026-toxsyn} \\
AssinRTE & Pair classification & \texttt{nilc-nlp/assin} & cc-by-4.0 & 2{,}000 & 2 classes & \cite{fonseca-etal-2016-assin} \\
InferBR & Pair classification & \texttt{hapaxlegomenon/InferBR} & mit & 1{,}705 & 2 classes & \cite{bencke-etal-2024-inferbr} \\
AssinSTS & STS & \texttt{nilc-nlp/assin} & cc-by-4.0 & 2{,}000 & continuous & \cite{fonseca-etal-2016-assin} \\
Assin2STS & STS & \texttt{nilc-nlp/assin2} & cc-by-4.0 & 2{,}448 & continuous & \cite{real2020assin} \\
MedPTClustering & Clustering & \texttt{AKCIT/MedPT} & cc-by-4.0 & 600 & 12 clusters & \cite{farber-etal-2026-medpt} \\
WikipediaPTCategoriesClusteringP2P & Clustering & \texttt{mteb-br/wikipedia-categories} & cc-by-sa-3.0 & 2{,}873 & 15 clusters & \cite{mtebpt2026wikicat} \\
JurisTCUClusteringP2P & Clustering & \texttt{mteb-br/juristcu-clustering} & cc-by-4.0 & 4{,}469 & 10 clusters & \cite{juristcu2026} \\
SciELOClusteringP2P & Clustering & \texttt{mteb-br/scielo-clustering} & cc-by-4.0 & 4{,}000 & 8 clusters & \cite{scielo_abstracts} \\
StackoverflowPtClustering & Clustering & \texttt{mteb-br/stackoverflow-clustering} & cc-by-sa-4.0 & 1{,}478 & 10 clusters & \cite{ptstackoverflow} \\
MedPTRetrieval & Retrieval & \texttt{AKCIT/MedPT} & cc-by-4.0 & 500 & 500 docs & \cite{farber-etal-2026-medpt} \\
FaQuADIR & Retrieval & \texttt{mteb-br/faquad-ir} & cc-by-4.0 & 900 & 244 docs & \cite{sayama-etal-2019-faquad} \\
Quati & Retrieval & \texttt{mteb-br/quati-50k} & cc-by-4.0 & 50 & 50{,}000 docs & \cite{bueno-etal-2024-quati} \\
FaqBacenRetrieval & Retrieval & \texttt{mteb-br/faq-bacen} & apache-2.0 & 373 & 1{,}673 docs & \cite{faq_bacen} \\
JurisTCU & Retrieval & \texttt{LeandroRibeiro/JurisTCU} & cc-by-4.0 & 150 & 16{,}045 docs & \cite{juristcu2026} \\
BRTaxQAR & Retrieval & \texttt{unicamp-dl/BR-TaxQA-R} & cc-by-4.0 & 598 & 478 docs & \cite{junior-etal-2025-brtaxqa} \\
QuatiReranking & Reranking & \texttt{mteb-br/quati-reranking} & cc-by-4.0 & 50 & 5{,}991 candidates & \cite{bueno-etal-2024-quati} \\
JurisTCUReranking & Reranking & \texttt{mteb-br/juristcu-reranking} & cc-by-4.0 & 150 & 7{,}429 candidates & \cite{juristcu2026} \\
PortuLexRRIP & Classification & \texttt{eduagarcia/PortuLex\_benchmark} & cc-by-4.0 & 1{,}474 & 8 classes & \cite{portulex_benchmark} \\
BrighterEmotionMultilabelClassification & Multilabel class. & \begin{tabular}[t]{@{}l@{}}\texttt{brighter-dataset/}\\\texttt{BRIGHTER-emotion-categories}\end{tabular} & cc-by-4.0 & 4{,}452 & 7 labels & \cite{brighter2025} \\
\bottomrule
\end{tabular}%
}
\end{table*}

\paragraph{Tasks.} Classification covers content moderation, fact-checking, and legal text: HateBR \cite{vargas-etal-2022-hatebr} (Instagram hate speech), ToxSynPT \cite{brito-etal-2026-toxsyn} (synthetic toxicity), FactckBr \cite{moreno2019factckbr} (fact-check classification), and PortuLexRRIP \cite{portulex_benchmark} (8-way legal rhetorical-role identification); BrighterEmotion \cite{brighter2025} adds multilabel emotion classification. Pair classification and STS draw on the ASSIN families: AssinRTE \cite{fonseca-etal-2016-assin} and InferBR \cite{bencke-etal-2024-inferbr} (entailment), and AssinSTS \cite{fonseca-etal-2016-assin} and Assin2STS \cite{real2020assin} (similarity). Clustering spans medical (MedPTClustering \cite{farber-etal-2026-medpt}), encyclopedic (the Wikipedia-category task introduced here), legal (JurisTCUClustering \cite{juristcu2026}), scientific (SciELOClustering \cite{scielo_abstracts}), and technical (StackoverflowClustering \cite{ptstackoverflow}). Retrieval spans open-domain (Quati \cite{bueno-etal-2024-quati}), legal (JurisTCU \cite{juristcu2026}), tax (BRTaxQA \cite{junior-etal-2025-brtaxqa}), banking FAQ (FaqBacen \cite{faq_bacen}), factual QA (FaQuADIR \cite{sayama-etal-2019-faquad}), and medical (MedPTRetrieval \cite{farber-etal-2026-medpt}); reranking reformulates the Quati and JurisTCU corpora as top-$k$ re-ranking. We report each model's score as the unweighted mean of its primary metric across the 22 headline tasks, hereafter the \emph{22-task mean} (written Mean$_{22}$ in tables), following the averaging convention of MTEB.

\paragraph{Exclusions.} Several candidate datasets are kept out of the headline suite. mMARCO-PT and mkqa-PT are machine translations of English benchmarks and fail the native-source filter. A second, empirical hazard reinforces this. Because MS MARCO is ubiquitous in retrieval-model training and mMARCO is its translation, scores on translated mMARCO-PT partly reflect prior exposure to that corpus rather than embedding quality, rising monotonically with a model's MS MARCO training (Appendix~\ref{app:mmarco}). The native retrieval tasks it displaces give a less train-on-test measurement of the same capability. We further exclude a clustering set that is degenerate (every model scores 1.000, as each cluster is a distinct news event with non-overlapping vocabulary, leaving no ranking signal), a news-clustering set that mirrors copyrighted content under no permissive license, and a classification variant with unclear provenance and an undeclared license. Beyond these named cases, the headline suite is the survivor of a wider screen. We tested many more Brazilian-Portuguese candidate tasks than we keep, dropping them by construction for degeneracy, low discrimination, unclear quality, or licensing. One filter was the per-task discrimination analysis of \S\ref{sec:irt}, which removes tasks at or below the random-baseline floor of \S\ref{sec:results}. The discriminations reported in \S\ref{sec:irt} are therefore those of the survivors, not of the candidate pool, and are biased upward; we report them to document the screen, not to claim that every retained task discriminates well. The final discrimination cut leaves the ranking essentially unchanged (Spearman $0.99$ between the 22-task means before and after it).

\begin{figure}[t]
\centering
\includegraphics[width=\columnwidth]{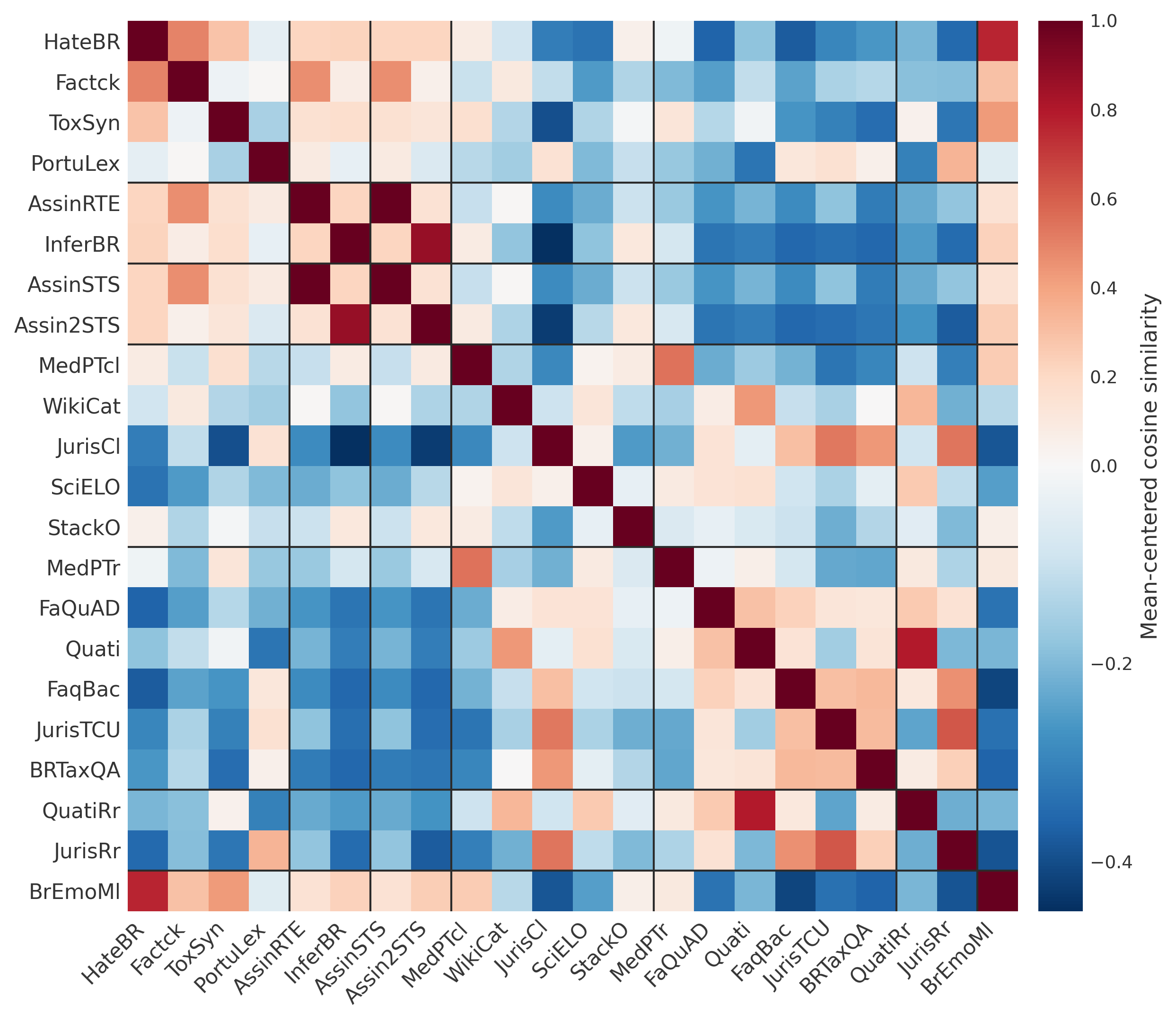}
\caption{\textbf{Corpus content overlap.} Mean-centered pairwise cosine similarity between per-task mean embeddings (100 documents per task), averaged over three architecturally diverse models (multilingual-E5-base, Qwen3-Embedding-0.6B, Serafim-100m). Within-category similarity exceeds across-category (values span $[-0.45, 1.00]$); the lone near-duplicate ($1.00$) is AssinRTE versus AssinSTS, which share a source corpus.}
\label{fig:inter_dataset_similarity}
\end{figure}

\paragraph{Diversity diagnostic.} To rule out accidental near-duplication, we follow BEIR \cite{thakur-etal-2021-beir} (also used by MTEB-NL \cite{banar-etal-2025-mteb-nl} and FaMTEB \cite{zinvandi-etal-2025-famteb}): we embed 100 documents per task and correlate the per-task mean embeddings. So that the result reflects corpus content rather than a single encoder's idiosyncrasy, we average over a panel of three architecturally diverse models: a multilingual encoder (multilingual-E5-base), a decoder-LLM embedder (Qwen3-Embedding-0.6B), and a Portuguese-specific encoder (Serafim-100m); the per-model similarity matrices agree closely (mean pairwise Spearman $\rho = 0.82$). Raw cosine is inflated because embedding vectors tend to share a common direction (anisotropy), which compresses every pair into a narrow high-similarity band, so we mean-center the task embeddings before correlating. After centering, tasks are on average nearly unrelated (mean pairwise similarity $-0.04$), and tasks in the same category are more alike than tasks across categories ($+0.04$ versus $-0.06$); the single $1.00$, AssinRTE--AssinSTS, which share the ASSIN corpus and differ only in label, stands out as the lone genuine near-duplicate (Figure~\ref{fig:inter_dataset_similarity}). The inter-task ranking-agreement analysis of \S\ref{sec:irt} reaches the same conclusion from model rankings rather than corpus content.

\section{Model Coverage}\label{sec:models}

We evaluate 93 models spanning three orders of magnitude in size, from 23M (all-MiniLM-L6-v2) to 27B (harrier-oss-v1-27b) parameters, grouped into five families; full per-model attributes (parameters, embedding dimension, context length) are given in Table~\ref{tab:model_catalog} and released with the results.

\emph{Decoder-LLM embedders} (32) adapt autoregressive language models as embedding backbones through contrastive fine-tuning: the Qwen3-Embedding \cite{qwen3-team-2024-qwen3}, Octen \cite{octen-2024}, F2LLM-v2 \cite{f2llm-2025}, harrier-oss \cite{harrier-2024}, SFR-Embedding \cite{sfr-2024}, and gte-Qwen2 \cite{li-etal-2023-gte} series, Linq-Embed-Mistral \cite{linq-2024}, KaLM-Embedding-Gemma3-12B \cite{kalm-2024}, Llama-Embed-Nemotron-8B \cite{nvidia-2024-llamanemotron}, BidirLM \cite{bidirlm-2024}, the BOOM checkpoint, and the Brazilian-legal-tuned jua models \cite{pereira-etal-2026-jua}, among others.
\emph{Multilingual encoders} (12) are from-scratch encoder-only models: the multilingual-E5 \cite{wang-etal-2024-e5}, paraphrase-multilingual \cite{reimers-gurevych-2019-sbert}, granite-embedding \cite{ibm-2024-granite}, and Jina-v5 \cite{jina-2024-v5} families, plus BGE-M3 \cite{chen-etal-2024-bge-m3} and Snowflake-arctic-embed-l-v2.0 \cite{snowflake-2024-arctic}.
\emph{Proprietary APIs} (20): Google Gemini-Embedding (001 and -2) and text-embedding-005, OpenAI text-embedding-3 (small and large), the Voyage 3, 3.5, and 4 generations (with their context, finance, and law variants), Mistral-embed and Codestral-embed, and Amazon Titan-Embed-Text-v2, each via its official endpoint at list price (\S\ref{sec:cost}). We evaluate every commercial embedding endpoint we could reach through a public API during the evaluation window, including Voyage's domain-specialized and context variants; coverage is consequently broader for vendors that expose more endpoints.
\emph{Portuguese-specific encoders} (16): the PORTULAN Serafim \cite{rodrigues-etal-2025-serafim} and Albertina \cite{portulan-2024-albertina} families, BERTimbau \cite{souza-etal-2020-bertimbau}, the Legal-BERTimbau and stjiris-legal-BERT variants \cite{stjiris-2024-legalbert}, and the MedLink biomedical bi-encoder.
\emph{Sentence-Transformer baselines} (13) anchor the low end: LaBSE \cite{feng-etal-2022-labse}, all-mpnet-base-v2 and all-MiniLM \cite{reimers-gurevych-2019-sbert}, embeddinggemma-300m \cite{google-2024-embeddinggemma}, GTE-small \cite{li-etal-2023-gte}, PIXIE-Rune-v1.0 \cite{pixie-2024}, and PwC-Embedding \cite{pwc-2024-embedding}, alongside the GIST, BGE-small, E5-small, and mxbai-embed-large public checkpoints; every model's identifier and attributes appear in Table~\ref{tab:model_catalog}.

\begin{table*}[t]
\centering
\caption{Metadata for the 93 evaluated models (read top-to-bottom, left panel then right). Ctx is the operative context length; Type: O (open) or C (closed); price is \$/1M input tokens (closed only).}
\label{tab:model_catalog}
\scriptsize
\resizebox{\textwidth}{!}{%
\begin{tabular}{lrrrcr@{\hspace{1.6em}}lrrrcr}
\toprule
Model & Params & Dim & Ctx & Type & \$/1M & Model & Params & Dim & Ctx & Type & \$/1M \\
\midrule
harrier-oss-v1-27b & 27.0B & 5376 & 131072 & O & --- & bert-large-portuguese-cased-legal-tsdae-gpl-nli-sts-MetaKD-v0 & 335M & 1024 & 512 & O & --- \\
F2LLM-v2-14B & 14.0B & 5120 & 40960 & O & --- & F2LLM-v2-330M & 334M & 896 & 40960 & O & --- \\
KaLM-Embedding-Gemma3-12B-2511 & 11.8B & 3840 & 32768 & O & --- & granite-embedding-311m-multilingual-r2 & 312M & 768 & 8192 & O & --- \\
Octen-Embedding-8B & 7.6B & 4096 & 32768 & O & --- & embeddinggemma-300m & 308M & 768 & 2048 & O & --- \\
Qwen3-Embedding-8B & 7.6B & 4096 & 32768 & O & --- & multilingual-e5-base & 278M & 768 & 512 & O & --- \\
F2LLM-v2-8B & 7.6B & 4096 & 40960 & O & --- & paraphrase-multilingual-mpnet-base-v2 & 278M & 768 & 512 & O & --- \\
llama-embed-nemotron-8b & 7.5B & 4096 & 32768 & O & --- & harrier-oss-v1-270m & 268M & 640 & 32768 & O & --- \\
gte-Qwen2-7B-instruct & 7.1B & 3584 & 32768 & O & --- & F2LLM-v2-160M & 159M & 640 & 40960 & O & --- \\
Linq-Embed-Mistral & 7.1B & 4096 & 32768 & O & --- & multilingual-e5-small & 118M & 384 & 512 & O & --- \\
SFR-Embedding-2\_R & 7.1B & 4096 & 32768 & O & --- & paraphrase-multilingual-MiniLM-L12-v2 & 118M & 384 & 512 & O & --- \\
SFR-Embedding-Mistral & 7.1B & 4096 & 32768 & O & --- & medlink-bi-encoder & 110M & 768 & 512 & O & --- \\
e5-mistral-7b-instruct & 7.1B & 4096 & 32768 & O & --- & bert-base-portuguese-cased & 110M & 768 & 512 & O & --- \\
BOOM\_4B\_v1 & 4.0B & 2560 & 32768 & O & --- & legal-bert-pt-br & 110M & 768 & 512 & O & --- \\
Qwen3-Embedding-4B & 4.0B & 2560 & 32768 & O & --- & all-mpnet-base-v2 & 109M & 768 & 384 & O & --- \\
F2LLM-v2-4B & 4.0B & 2560 & 40960 & O & --- & granite-embedding-107m-multilingual & 107M & 384 & 512 & O & --- \\
jua-4B-legal-only & 4.0B & 2560 & 40960 & O & --- & serafim-100m-portuguese-pt-sentence-encoder & 100M & 768 & 512 & O & --- \\
jua-4B-mixed & 4.0B & 2560 & 8192 & O & --- & serafim-100m-portuguese-pt-sentence-encoder-ir & 100M & 768 & 128 & O & --- \\
BidirLM-1.7B-Embedding & 1.7B & 2048 & 512 & O & --- & granite-embedding-97m-multilingual-r2 & 97M & 384 & 8192 & O & --- \\
F2LLM-v2-1.7B & 1.7B & 2048 & 40960 & O & --- & F2LLM-v2-80M & 80M & 320 & 40960 & O & --- \\
gte-Qwen2-1.5B-instruct & 1.5B & 1536 & 32768 & O & --- & bge-small-en-v1.5 & 33M & 512 & 512 & O & --- \\
BidirLM-1B-Embedding & 1000M & 1152 & 512 & O & --- & e5-small-v2 & 33M & 384 & 512 & O & --- \\
albertina-900m-portuguese-ptbr-encoder & 900M & 1536 & 512 & O & --- & all-MiniLM-L12-v2 & 33M & 384 & 256 & O & --- \\
serafim-900m-portuguese-pt-sentence-encoder & 900M & 1536 & 512 & O & --- & gte-small & 33M & 384 & 512 & O & --- \\
serafim-900m-portuguese-pt-sentence-encoder-ir & 900M & 1536 & 128 & O & --- & Ivysaur & 23M & 384 & 512 & O & --- \\
Qwen3-Embedding-0.6B-jua-V2 & 600M & 1024 & 32768 & O & --- & GIST-all-MiniLM-L6-v2 & 23M & 384 & 512 & O & --- \\
BidirLM-0.6B-Embedding & 596M & 1024 & 512 & O & --- & all-MiniLM-L6-v2 & 23M & 384 & 256 & O & --- \\
Octen-Embedding-0.6B & 596M & 1024 & 32768 & O & --- & titan-embed-text-v2 & --- & 1024 & 8192 & C & 0.020 \\
Qwen3-Embedding-0.6B & 596M & 1024 & 32768 & O & --- & gemini-embedding-001 & --- & 3072 & 2048 & C & 0.150 \\
F2LLM-0.6B & 596M & 1024 & 8192 & O & --- & gemini-embedding-2 & --- & 3072 & 8192 & C & 0.200 \\
F2LLM-v2-0.6B & 596M & 1024 & 40960 & O & --- & text-embedding-005 & --- & 768 & 2048 & C & 0.030 \\
jina-embeddings-v5-text-small & 596M & 1024 & 32768 & O & --- & codestral-embed & --- & 1536 & 8192 & C & 0.150 \\
harrier-oss-v1-0.6b & 596M & 1024 & 32768 & O & --- & mistral-embed & --- & 1024 & 8192 & C & 0.100 \\
bge-m3 & 568M & 1024 & 8192 & O & --- & text-embedding-3-large & --- & 3072 & 8192 & C & 0.130 \\
snowflake-arctic-embed-l-v2.0 & 568M & 1024 & 8192 & O & --- & text-embedding-3-small & --- & 1536 & 8192 & C & 0.020 \\
PIXIE-Rune-v1.0 & 568M & 1024 & 6144 & O & --- & voyage-3 & --- & 1024 & 32000 & C & 0.060 \\
PwC-Embedding\_expr & 560M & 1024 & 512 & O & --- & voyage-3-large & --- & 1024 & 32000 & C & 0.180 \\
multilingual-e5-large & 560M & 1024 & 512 & O & --- & voyage-3-lite & --- & 512 & 32000 & C & 0.020 \\
multilingual-e5-large-instruct & 560M & 1024 & 512 & O & --- & voyage-3.5 & --- & 1024 & 32000 & C & 0.060 \\
LaBSE & 471M & 768 & 512 & O & --- & voyage-3.5-lite & --- & 1024 & 32000 & C & 0.020 \\
serafim-335m-portuguese-pt-sentence-encoder & 335M & 1024 & 512 & O & --- & voyage-4 & --- & 1024 & 32000 & C & 0.060 \\
serafim-335m-portuguese-pt-sentence-encoder-ir & 335M & 1024 & 128 & O & --- & voyage-4-large & --- & 1024 & 32000 & C & 0.120 \\
mxbai-embed-large-v1 & 335M & 1024 & 512 & O & --- & voyage-4-lite & --- & 1024 & 32000 & C & 0.020 \\
bert-large-portuguese-cased & 335M & 1024 & 512 & O & --- & voyage-context-3 & --- & 1024 & 32000 & C & 0.180 \\
Legal-BERTimbau-sts-large & 335M & 1024 & 512 & O & --- & voyage-context-4 & --- & 1024 & 32000 & C & 0.120 \\
Legal-BERTimbau-sts-large-ma-v3 & 335M & 1024 & 512 & O & --- & voyage-finance-2 & --- & 1024 & 32000 & C & 0.120 \\
bert-large-portuguese-cased-legal-mlm-mkd-nli-sts-v1 & 335M & 1024 & 512 & O & --- & voyage-law-2 & --- & 1024 & 16000 & C & 0.120 \\
bert-large-portuguese-cased-legal-mlm-sts-v1.0 & 335M & 1024 & 512 & O & --- &  &  &  &  &  &  \\
\bottomrule
\end{tabular}%
}
\end{table*}

\section{Evaluation Protocol}\label{sec:protocol}

We adopt the per-category metrics of MTEB \cite{muennighoff-etal-2023-mteb}, inherited unchanged by every language-specific extension, so MTEB-BR scores remain comparable to global MTEB up to the language-domain shift: classification is accuracy from a logistic-regression probe over 10 train/test resamples; multilabel classification is subset accuracy (the fraction of instances with every label predicted correctly); pair classification is held-out accuracy; STS is the Spearman rank correlation (1 means identical order) between cosine similarity and human judgment; clustering is V-measure (a clustering-quality score from 0 to 1) over 10 $k$-means runs; retrieval is nDCG@10 (a ranking score from 0 to 1 that rewards placing relevant documents in the top 10); and reranking is mean average precision (MAP, precision averaged along the ranking). For every (task, model) pair we publish the headline score, the per-instance score distribution where it is well-defined (per-query nDCG@10, per-experiment V-measure, per-resample accuracy), and the dataset revision SHA, all as Parquet in the results repository.

\paragraph{The statistical layer.} For every headline comparison we report a set of complementary quantities, each answering a plain question about the ranking.
\emph{Bootstrap confidence intervals} ask how much a score would move if the benchmark were re-run on a resampled task suite: a 95\% percentile-bootstrap CI from 10{,}000 resamples of the underlying score distribution. (The two pair-classification and two STS tasks lack per-instance distributions in \texttt{mteb} 2.12, so we bracket them with standard analytic confidence intervals (Fisher-$z$ for the STS correlations, Wilson for the pair accuracies), tight for the leading models at $\pm 0.013$--$0.016$.)
\emph{Paired significance} asks whether one model reliably beats another on the same tasks, by resampling the 22 tasks with replacement (10{,}000 resamples) and recomputing the paired difference. From the same resamples we read two plain summaries used in \S\ref{sec:results}: the \emph{probability that a model is best} (the fraction of resamples in which it attains the top mean) and \emph{pairwise resolution} (the fraction in which one model outranks another); models that no pair separates at $95\%$ form a \emph{tier}.
\emph{Item Response Theory} treats the score matrix as a test: a task's \emph{discrimination} $a_t$ measures how sharply it separates models of similar ability, and a model's \emph{ability} $\theta$ is a skill estimate that weights each task by that discrimination (\S\ref{sec:irt}); retrieval tasks score $a_t \in [1.49, 2.05]$ and clustering tasks $a_t \in [0.67, 1.22]$.
\emph{Borda count} is a preference-vote ranking, each task casting a vote, robust to the odd outlier task \cite{colombo-etal-2022-bordatte} and agreeing with the 22-task-mean ranking at Kendall $\tau = 0.91$ (a rank-agreement measure where 1 is identical order).

\section{Headline Results}\label{sec:results}

Table~\ref{tab:top15} ranks the top 15 of 93 models by the 22-task mean, with 95\% bootstrap confidence intervals. Across the full panel, the 22-task mean spans $0.248$ (paraphrase-multilingual-MiniLM-L12-v2) to $0.682$ (gemini-embedding-001), against a random-embedding baseline whose 22-task mean is $0.18$, with per-task chance levels that vary sharply by category (near zero for retrieval, higher for classification and clustering; Table~\ref{tab:full_matrix}); the complete 93-model ranking is on the interactive leaderboard. The leaders mix closed commercial APIs and large open-weight decoder-LLM embedders.

\begin{table*}[t]
\centering
\caption{Top-15 models on MTEB-BR by the 22-task mean, with 95\% bootstrap intervals (score and rank) from 10{,}000 resamples of the 22 headline tasks. The top six form one converged frontier tier, their pairwise order unresolved (\S\ref{sec:results}). Score intervals are marginal per-model resamples, not the paired significance test (Appendix~\ref{app:significance}).}
\label{tab:top15}
\small
\begin{tabular}{rlcrrcc}
\toprule
Rank & Model & Type & Params & Mean$_{22}$ & Score 95\% CI & Rank 95\% CI \\
\midrule
1 & gemini-embedding-001 & closed & --- & 0.6820 & [0.6057, 0.7533] & [1, 5] \\
2 & Qwen3-Embedding-8B & open & 7.6B & 0.6704 & [0.5902, 0.7437] & [1, 12] \\
3 & KaLM-Embedding-Gemma3-12B-2511 & open & 11.8B & 0.6701 & [0.5864, 0.7490] & [1, 11] \\
4 & voyage-context-4 & closed & --- & 0.6676 & [0.5924, 0.7388] & [1, 16] \\
5 & Octen-Embedding-8B & open & 7.6B & 0.6674 & [0.5857, 0.7416] & [2, 12] \\
6 & Qwen3-Embedding-4B & open & 4.0B & 0.6621 & [0.5833, 0.7344] & [1, 23] \\
7 & voyage-context-3 & closed & --- & 0.6571 & [0.5816, 0.7281] & [3, 21] \\
8 & voyage-3-large & closed & --- & 0.6552 & [0.5811, 0.7246] & [3, 24] \\
9 & voyage-4-large & closed & --- & 0.6532 & [0.5791, 0.7244] & [3, 26] \\
10 & SFR-Embedding-Mistral & open & 7.1B & 0.6523 & [0.5689, 0.7289] & [4, 27] \\
11 & BidirLM-1.7B-Embedding & open & 1.7B & 0.6513 & [0.5671, 0.7293] & [3, 31] \\
12 & BOOM\_4B\_v1 & open & 4.0B & 0.6503 & [0.5704, 0.7232] & [6, 24] \\
13 & embeddinggemma-300m & open & 308M & 0.6490 & [0.5723, 0.7207] & [6, 26] \\
14 & codestral-embed & closed & --- & 0.6486 & [0.5647, 0.7265] & [5, 30] \\
15 & Linq-Embed-Mistral & open & 7.1B & 0.6473 & [0.5605, 0.7265] & [6, 33] \\
\bottomrule
\end{tabular}
\end{table*}

\paragraph{Resolution across the field.} A useful benchmark should order the models it is given, and MTEB-BR does so across most of the field. Of all 4{,}278 model pairs, $78.7\%$ are cleanly separated (one model's 22-task mean beats the other's in over $95\%$ of task-bootstrap resamples); adjacent-rank pairs are the hard case, and only about $3\%$ of them resolve, so what the benchmark orders cleanly is distant pairs, not near-neighbours. Grouping models it cannot separate yields roughly a dozen tiers (8 to 15 as the resolution threshold moves from $0.99$ to $0.90$), and the leader, gemini-embedding-001, stands above 87 of the other 92. The 22-task mean spans 43 points end to end, with tight rank intervals through the mid-field (Table~\ref{tab:top15}).

Only at the very top do the models stop separating. A typical top-six pair differs by about $0.005$ in mean while individual tasks vary by roughly $0.05$, so separating adjacent frontier ranks would take hundreds of tasks. That is an upper bound rather than a goal, since a gap so small may itself be noise; finer resolution at the top is better sought through harder or more discriminating tasks (\S\ref{sec:irt}) than through more tasks of the present difficulty.

\paragraph{The converged frontier tier.} Within that cluster the leaderboard converges. The six leaders' means fall within $0.020$ of one another, and the paired bootstrap does not resolve their order. gemini is nonetheless the leader: it leads the unweighted mean and every robust aggregation we tried, a per-task z-score mean, a discrimination-weighted mean, and Borda count, and across task-resampled suites it holds first place in $66\%$ of draws against $13\%$ for the next. The one summary that ranks it second is a least-squares Item Response Theory ability fit (\S\ref{sec:irt}), and only by under $0.001$ standard deviations. This convergence is a property of these frontier models, not of the benchmark, which separates the cluster decisively from the rest of the field; within it the choice turns on cost (\S\ref{sec:cost}), license, latency, and context length rather than score.

\paragraph{Robust to aggregation and within-task noise.} The frontier convergence is not an artefact of how scores are combined. First, the paired bootstrap resamples the 22 tasks, treating each task's score as fixed. Its interval therefore measures how much the ranking depends on which tasks are in the suite, not sampling error within a task. That within-task noise is small: even the smallest task (Quati, 50 queries) has a per-query standard error of $0.035$, which contributes only $0.0016$ to the 22-task mean, well below the $0.039$ spread the resample already captures across tasks. A paired per-query two-level bootstrap confirms this: resampling within tasks barely moves the top-tier interval. Second, the ranking is robust to the aggregation weight: a category-balanced mean (equal weight to each of the seven categories rather than to each task) reorders the panel at Kendall $\tau = 0.91$, eight of the top ten unchanged, so the unweighted-mean convention (inherited from MTEB) is not driving the result despite the $100\times$ spread in evaluation-set sizes.

Qwen3-Embedding-8B sits inside this cluster at rank two, alongside KaLM-Embedding-Gemma3-12B at three. It is open-weight (Apache-2.0) and self-hostable \cite{qwen3-team-2024-qwen3}, yet the median of the other 92 models beats it on only one of the 22 tasks, and the closed APIs that lead retrieval do so by margins inside the bootstrap band. An open-weight model thus sits among the closed leaders on the cost--quality frontier. At the other end of the scale axis, the 308M embeddinggemma-300m ranks 13th of 93, within $0.033$ of the leader: near-frontier quality at edge-deployable size.

\paragraph{Retrieval as the dominant separator.} Per-category means show retrieval spreading models most widely: it carries the highest per-task discrimination and the largest cross-model score variance of any category (\S\ref{sec:irt}, Table~\ref{tab:irt}). No single model leads all of retrieval, classification, and clustering, and the clustering tasks separate models weakly (the lowest-discriminating tasks are split between clustering and two classification probes, \S\ref{sec:irt}). We take up the discrimination-theoretic basis for this in \S\ref{sec:irt}.

\paragraph{Portuguese-specific encoders and retrieval.} Brazilian-Portuguese-specific encoders trail general multilingual encoders on the retrieval and reranking tasks, but the gap reflects training objective and domain, not language specialization. The 16 Portuguese-specific encoders average $0.331$ across these eight tasks against $0.517$ for the 12 general multilingual encoders, a 0.186-point gap (Welch's $t$-test $t = -3.45$, $p = 0.003$; a rank-based Mann--Whitney test agrees, $p = 0.001$). That comparison is confounded by training objective: the Portuguese-specific group includes four masked-language-model encoders never fine-tuned for retrieval (BERTimbau and Albertina, mean $0.258$) alongside legal-domain encoders specialized away from open-domain search. The general-purpose, retrieval-tuned Serafim encoders, by contrast, show no detectable gap from the multilingual group ($0.516$ versus $0.517$, $\Delta \approx 0$, $p = 0.99$); with only three IR-tuned Serafim variants the sample is too small to prove they are equal, so we read this as finding no gap rather than as demonstrated equivalence, but it still places the difference in training objective and domain rather than language. Scale and breadth of contrastive training, not language-specific pretraining, plausibly govern retrieval quality here \cite{wang-etal-2024-e5,chen-etal-2024-bge-m3}.

\paragraph{The classification mirror image.} The same masked-language-model encoders that trail on retrieval rank among the leading models on classification, where MTEB fits a logistic-regression probe on frozen embeddings and so rewards linearly separable features rather than metric geometry. Averaged over the four classification tasks, the 335M BERTimbau-large \cite{souza-etal-2020-bertimbau} ranks 3rd of 93 (second among open-weight models, behind only the 27B harrier-oss) and the 110M BERTimbau-base 9th, both ahead of every open decoder-LLM embedder except harrier and of most closed APIs, though neither was fine-tuned for embeddings. An encoder can thus carry probeable class signal without the metric geometry open-domain retrieval demands: the same training-objective account read from the other side.

\section{Cross-Leaderboard Agreement}\label{sec:cross_leaderboard}

Does a single-language benchmark add information beyond the existing multilingual leaderboards? If MTEB-BR rankings tracked the global HuggingFace MTEB multilingual leaderboard closely, a practitioner could treat the latter as a proxy and this benchmark would be largely presentational. We test the proxy hypothesis directly. Matching the 55 headline-panel models that appear on both leaderboards (HuggingFace snapshot of 2026-06-29; models matched by normalized name), the Spearman correlation between the HuggingFace MTEB multilingual rank and the MTEB-BR 22-task-mean rank is $\rho = 0.754$ ($n = 55$), with a model-resampling 95\% bootstrap interval of $[0.54, 0.89]$; the Pearson correlation is $0.766$. The multilingual ranking carries real information about Portuguese performance, but even the upper end of that interval falls short of a usable proxy and roughly 40\% of the variance is unexplained ($R^2 \approx 0.59$); the live board is a drifting, differently-composed reference, so the correlation should be read as moderate, not exact.

The agreement is uneven across categories: STS transfers most cleanly ($\rho = 0.80$), followed by reranking ($0.75$) and retrieval ($0.74$), with pair classification ($0.73$) and classification ($0.68$) in the middle and clustering least ($0.62$). Even STS, the highest-agreement category, stays near $\rho = 0.80$, so the multilingual ranking leaves substantial residual variance in every category, and language-specific extensions are not redundant restatements of it.

\begin{figure}[t]
\centering
\includegraphics[width=\columnwidth]{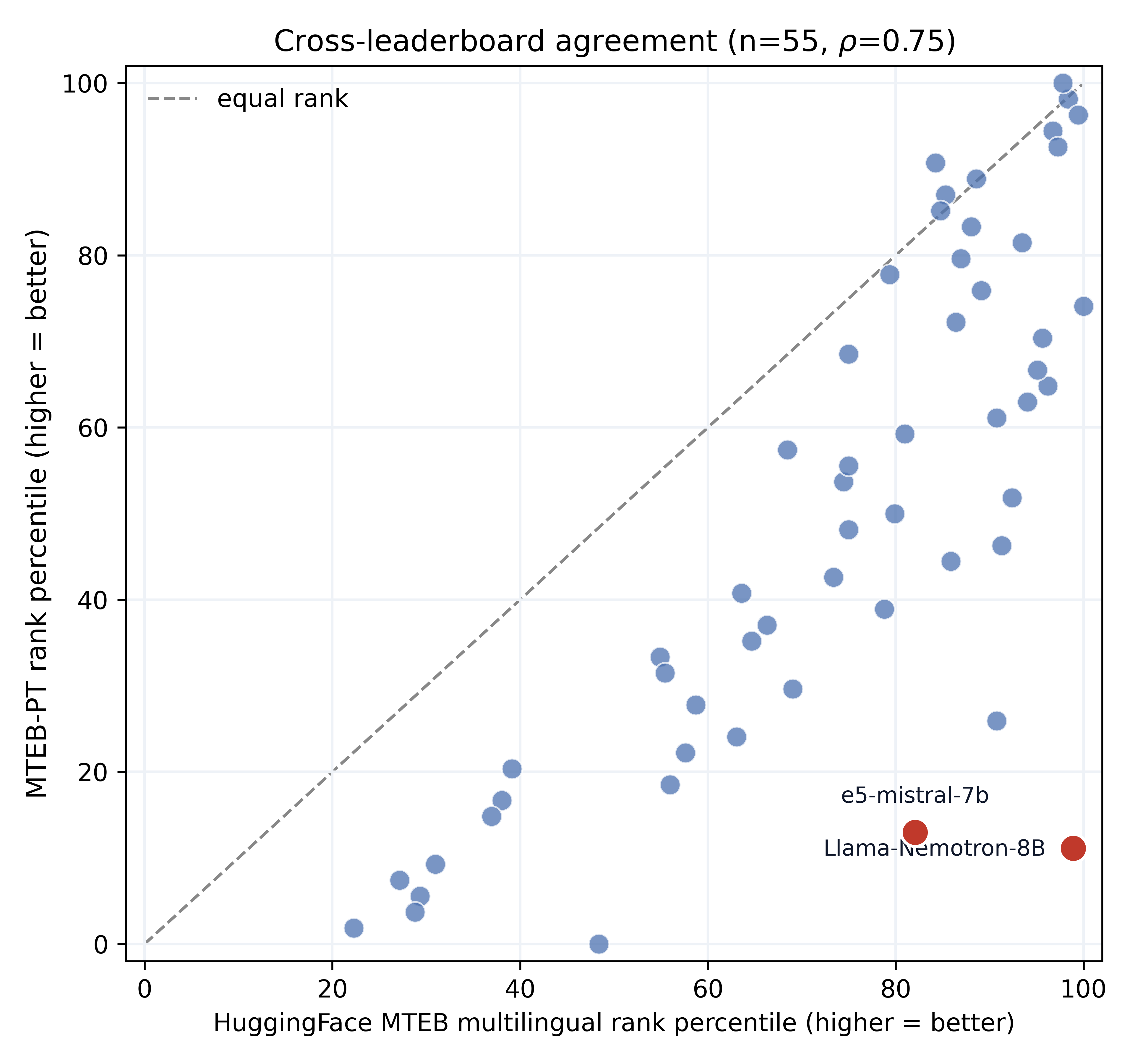}
\caption{HuggingFace MTEB multilingual versus MTEB-BR rank, expressed as within-leaderboard percentile, for the 55 matched models. Most models sit near the diagonal; the two largest high-multilingual yet low-Portuguese divergences (Llama-Embed-Nemotron-8B and e5-mistral-7b-instruct) fall far below it, near the bottom of the shared set despite ranking high on the multilingual board.}
\label{fig:cross_scatter}
\end{figure}

The largest shift is Llama-Embed-Nemotron-8B \cite{nvidia-2024-llamanemotron} (Figure~\ref{fig:cross_scatter}), a multilingual model ranking 3rd of these 55 on the multilingual leaderboard yet 49th of them on MTEB-BR, the largest such divergence in the panel. The divergence concentrates in retrieval: on the six MTEB-BR retrieval tasks it falls far below the panel median, while on classification and clustering the gap is smaller but in the same direction. One plausible cause is the composition of the multilingual leaderboard, which is dominated by English-mined data and mMARCO with little native low-resource retrieval; a model strong there need not transfer to native Brazilian-Portuguese retrieval. Part of this divergence is moreover a designed consequence: because MTEB-BR excludes translated retrieval such as mMARCO by construction, a model strong on that style of data necessarily diverges here, so the low correlation partly reflects our task selection rather than a deficiency in the multilingual board.

A second high-multilingual model diverges the same way for a different reason: e5-mistral-7b-instruct diverges less sharply, mid-pack on the multilingual board (26th of these 55) but 48th on MTEB-BR, its English-only Mistral-7B base carrying little native Portuguese. Nor is the divergence specific to this one live snapshot. The published MMTEB headline \cite{enevoldsen-etal-2025-mmteb}, built from a different task pool, model set, and aggregation, ranks multilingual-E5-large-instruct first among its open models and reports that broad multilingual pretraining compensates for model size. Yet that model places 19th of all 93 on MTEB-BR, behind the decoder-LLM embedders that lead native Brazilian-Portuguese; a second, independently-constructed multilingual reference reorders the same way.

The multilingual leaderboard remains a useful first filter (a model outside the global top tier rarely leads here), but it cannot substitute for native evaluation. We do not claim the MTEB-BR ranking is closer to real deployment quality: proving that would need a separate downstream task, which is outside this study.

\section{Task Discrimination via Item Response Theory}\label{sec:irt}

A benchmark of twenty-two tasks raises a measurement question: which tasks actually separate models, and which contribute redundant or saturated signal? Item Response Theory (IRT), developed in psychometrics to estimate how sharply a test item distinguishes examinees of differing ability \cite{hambleton-etal-1991-irt}, answers this directly. We treat each task as an item and each model as an examinee and fit a two-parameter logistic (2-PL) response function,
\begin{equation}
\hat{s}_{mt} = \frac{1}{1 + \exp\!\big(-a_t\,(\theta_m - b_t)\big)},
\end{equation}
where $\hat{s}_{mt} \in (0,1)$ is the predicted normalized score of model $m$ on task $t$, recovering a discrimination $a_t$ (the slope of the task's response curve) and a difficulty $b_t$ for every task, and a latent ability $\theta_m$ for every model. IRT has been applied to NLP evaluation before (to scale item difficulty \cite{lalor-etal-2016-irt}, to select informative examples for cheaper evaluation \cite{polo-etal-2024-tinybenchmarks}, and to re-weight leaderboard examples \cite{rodriguez-etal-2021-evaluation}), but always at the \emph{instance} level, where items are individual test examples and the response is dichotomous. We apply it at two granularities: here at the \emph{task} level, as a benchmark-design diagnostic over the suite's twenty-two tasks, and at the canonical instance level in \S\ref{sec:irt_instance}. To our knowledge no benchmark in the MTEB family reports either.

Because benchmark scores are continuous in $[0,1]$ rather than dichotomous, we fit the 2-PL by least squares (gradient descent, 5{,}000 iterations, $\theta$ mean-centered for identifiability) on the 93-model panel, rather than by the standard dichotomous-response maximum likelihood. We report the recovered discriminations $a_t$ and difficulties $b_t$ in Table~\ref{tab:irt}; the model abilities $\theta_m$ reproduce the 22-task-mean ranking (Spearman $\rho = 0.99$).

\begin{table}[htbp]
\centering
\caption{Per-task IRT discrimination $a_t$ (95\% bootstrap CI, 200 resamples of the 93-model panel) and difficulty $b_t$ from the 2-PL fit, ranked by $a_t$. $\mu$,$\sigma$ are cross-model score mean and std. Retrieval dominates; clustering and the weakest classification tasks discriminate least.}
\label{tab:irt}
\small
\resizebox{\columnwidth}{!}{%
\begin{tabular}{lllrrr}
\toprule
Task & Category & $a_t$ [95\% CI] & $b_t$ & $\mu$ & $\sigma$ \\
\midrule
MedPTRetr & Retrieval & 2.048 [1.89, 2.17] & -0.233 & 0.665 & 0.197 \\
FaqBacen & Retrieval & 1.876 [1.72, 2.02] & -0.072 & 0.586 & 0.186 \\
Quati & Retrieval & 1.784 [1.62, 1.95] & +0.189 & 0.474 & 0.175 \\
FaQuADIR & Retrieval & 1.761 [1.51, 1.97] & -0.437 & 0.714 & 0.164 \\
BRTaxQAR & Retrieval & 1.574 [1.47, 1.66] & +0.890 & 0.241 & 0.121 \\
JurisTCU & Retrieval & 1.486 [1.29, 1.64] & +0.180 & 0.481 & 0.150 \\
QuatiRerank & Reranking & 1.430 [1.34, 1.50] & +0.119 & 0.504 & 0.133 \\
InferBR & Pair classification & 1.268 [1.07, 1.47] & -0.585 & 0.704 & 0.136 \\
SciELOClus & Clustering & 1.218 [0.98, 1.42] & -0.225 & 0.601 & 0.138 \\
ToxSyn & Classification & 1.205 [1.11, 1.32] & -1.203 & 0.821 & 0.075 \\
AssinRTE & Pair classification & 1.186 [1.10, 1.28] & -0.986 & 0.779 & 0.085 \\
Assin2STS & STS & 1.147 [1.02, 1.29] & -0.717 & 0.717 & 0.102 \\
HateBR & Classification & 1.132 [1.04, 1.24] & -1.082 & 0.788 & 0.068 \\
AssinSTS & STS & 1.101 [1.00, 1.23] & -0.749 & 0.716 & 0.092 \\
JurisTCURerank & Reranking & 1.052 [0.96, 1.13] & +0.225 & 0.475 & 0.097 \\
JurisClus & Clustering & 1.027 [0.94, 1.12] & +0.892 & 0.318 & 0.080 \\
BrEmoMulti & Multilabel & 0.993 [0.94, 1.07] & +1.203 & 0.262 & 0.034 \\
WikiCatClus & Clustering & 0.805 [0.67, 0.91] & -0.417 & 0.607 & 0.096 \\
MedPTClus & Clustering & 0.714 [0.60, 0.84] & -0.954 & 0.687 & 0.069 \\
StackOClus & Clustering & 0.672 [0.55, 0.81] & +0.343 & 0.463 & 0.073 \\
PortuLexRRI & Classification & 0.505 [0.44, 0.60] & +0.871 & 0.398 & 0.053 \\
Factck & Classification & 0.467 [0.37, 0.58] & +0.035 & 0.512 & 0.049 \\
\bottomrule
\end{tabular}%
}
\end{table}

Retrieval tasks discriminate models sharply; clustering and the weakest classification tasks do not. The six retrieval tasks occupy the top six positions in Table~\ref{tab:irt}, led by MedPTRetrieval ($a_t = 2.05$) and FaqBacen ($a_t = 1.88$). The two lowest of all are classification tasks: fact-check verification (FactckBr, $0.47$) and legal rhetorical-role identification (PortuLexRRIP, $0.51$), followed by three of the five clustering tasks ($a_t$ between $0.67$ and $0.81$). Averaged by category, classification is the lowest at $0.83$ ($47\%$ of retrieval's $1.75$), but that average is pulled down by those two probes while HateBR ($1.13$) and ToxSynPT ($1.21$) discriminate well; the more uniform clustering category averages $0.89$, and multilabel classification, STS, pair classification, and reranking fall in between (all near or above $1.0$; Table~\ref{tab:irt}). This has a practical reading: a fixed budget for separating similar-ability models is better spent on retrieval coverage than on clustering or the low-signal classification probes, which carry little of the ability-ranking signal at the headline level.

Bootstrap CIs on $a_t$ (resampling the 93-model panel 200 times, so they gauge stability under panel composition rather than classical estimation error; Table~\ref{tab:irt}) make the category-level separation robust (Quati at $1.78$ $[1.62, 1.95]$ does not overlap MedPTClustering at $0.71$ $[0.60, 0.84]$), while adjacent within-category ranks overlap, so the ordering is reliable by category, not as an exact per-task rank. Read the $a_t$ as a descriptive, panel-relative diagnostic, not a psychometric estimate. It is a least-squares 2-PL fit on continuous scores with a single latent ability, so its values are meaningful only in their ordering and ratios. As expected for a single-factor fit, they track the raw per-task score variance closely ($\rho = 0.88$, Table~\ref{tab:irt}). The task-level $a_t$ thus compactly restate which tasks spread models widely. The generative uses of IRT (efficient item selection, misfit as a data-quality signal) we reserve for the instance level (\S\ref{sec:irt_instance}), where the response is dichotomous.

\paragraph{Model ability.} Fitting the same 2-PL as an ability model, a latent $\theta_m$ per model estimated jointly with the task parameters, recovers the 22-task-mean ranking closely over the full panel (Kendall $\tau = 0.91$). At the very top the two disagree: this least-squares ability fit, which we read as descriptive rather than psychometric, places voyage-context-4 marginally above gemini, by under $0.001$ standard deviations, where the mean and the other aggregations place gemini first. The disagreement falls inside the converged frontier tier (\S\ref{sec:results}) and does not settle its order.

A second view of redundancy, complementary to the corpus-content diagnostic of \S\ref{sec:benchmark}, is direct task-to-task ranking agreement: it ignores corpus text and reads only the score matrix. Figure~\ref{fig:irt_spearman} (Appendix~\ref{app:spearman}) reports the Spearman rank correlation between every pair of tasks over the 93-model panel. The mean off-diagonal correlation is $\bar{\rho} = 0.53$, ranging from $-0.16$ (MedPTClustering vs.\ PortuLexRRIP, unrelated categories) to $0.97$ (Quati vs.\ QuatiReranking, the same corpus under two protocols). Within-category agreement ($\bar{\rho} = 0.67$) exceeds across-category agreement ($0.51$); the clustering tasks are the exception, agreeing among themselves at only $\rho = 0.47$, the lowest of any category and the same conclusion the IRT fit reaches by a different route. IRT asks which tasks separate models of similar ability; Spearman asks which tasks agree on the ordering. The two answer adjacent questions; since both are computed from the same score matrix, their agreement is expected rather than independent corroboration.

Finally, the headline ranking does not depend on the choice of aggregation. Across the 93 models, the Borda count (each task casting a preference vote with a tournament tie-break \cite{colombo-etal-2022-bordatte}) agrees with the simple per-task mean at Kendall $\tau = 0.91$; the two orderings differ only for models with high cross-task variance, which Borda penalizes and the mean does not, and never by more than a few positions.

Nor does it depend on the suite's shared-corpus task pairs. The ASSIN sentences seed both AssinRTE and AssinSTS, and the JurisTCU and Quati corpora each recur under multiple protocols (retrieval, reranking, and, for JurisTCU, clustering), so the 22-task mean treats some correlated signal as independent. Dropping the redundant reformulations (the two corpus-derived reranking tasks, the JurisTCU clustering task, and one of the two ASSIN tasks) leaves a ranking that agrees with the full 22-task mean at Kendall $\tau = 0.97$ (Spearman $0.997$, with $7$ of the top ten unchanged), and leaves both the top-cluster spread and the cross-leaderboard correlation ($\rho \approx 0.75$) intact. The suite carries redundant pairs, but the headline conclusions do not rest on counting them as independent tasks.

\subsection{Instance-level discrimination}\label{sec:irt_instance}

The task-level fit asks which task \emph{types} separate models. The same model applied \emph{within} a task (items are individual test instances, the response is whether a model answers each instance correctly) is the canonical dichotomous 2-PL of the IRT-for-NLP literature \cite{lalor-etal-2016-irt,polo-etal-2024-tinybenchmarks} and answers a sharper question: which \emph{instances} carry the discriminative signal. We fit it by penalized marginal maximum likelihood (EM with Gauss--Hermite quadrature; a weak $\log a \sim \mathcal{N}(0,1)$ prior regularizes low-variance items) on the four dichotomous tasks whose per-instance predictions are released (HateBR and ToxSynPT for classification, AssinRTE and InferBR for pair classification), over the 35--37 models with per-sample dumps. Recovered abilities reproduce per-task accuracy (Kendall $\tau = 0.74$--$0.96$). Two benchmark-relevant uses follow.

\paragraph{Efficient evaluation.} Ranking instances by Fisher information (a measure of how much a single test example reveals about a model's ability) and scoring models on only the most informative ones recovers the full-task ranking with far fewer instances than random sampling. Across the four tasks the $K = 50$ highest-information instances recover the full-task ranking at Kendall $\tau = 0.66$--$0.90$, versus $0.42$--$0.65$ for random selection of the same size; the margin is largest on the synthetic ToxSynPT ($0.90$ vs $0.50$, of $5{,}208$ instances), whose LM-generated items may be more template-separable, and smaller on the human-annotated tasks (HateBR $0.70$ vs $0.52$), so the latter is the conservative estimate. The advantage holds for $K = 25$--$200$ and survives cross-model validation: selecting items on one half of the models and ranking the disjoint other half still beats random ($\tau = 0.82$ vs $0.55$ on held-out models at $K = 50$, ToxSynPT). As a sanity check, selecting the \emph{least}-informative instances instead drives the recovered ranking negative ($\tau = -0.76$ at $K = 25$): informative items help and the least-informative ones actively hurt, which confirms the fit is finding real signal rather than noise \cite{polo-etal-2024-tinybenchmarks}.

\paragraph{A label-quality audit.} Instances whose correctness correlates \emph{negatively} with model ability (strong models systematically wrong, weak models right) are item-misfit flags, $11\%$ (ToxSynPT) to $30\%$ (AssinRTE) of active items. Inspecting the worst HateBR cases is suggestive rather than a re-annotation (confirming them would require a held-out human audit), but they consistently surface annotation-subjective boundary instances rather than random noise or memorized duplicates, yielding a ranked, auditable list of the most fit-anomalous instances per task; on near-ceiling tasks some of these reflect model misfit rather than mislabeling, so the list is a label-review screen, not a verdict.

This is a case study, not a benchmark-wide claim. Per-instance predictions are released for seven of the twenty-two tasks; we fit the four with binary instance-level correctness (classification and pair classification), leaving aside the two continuous STS tasks and one further classification task. The fit spans 35--37 of the 93 models, the efficiency result depends on a correctly regularized fit (an unregularized joint-ML fit is degenerate), and pair-task correctness is threshold-dependent. Full per-task results are in Table~\ref{tab:irt_instance}.

\section{Resource-Aware Model Selection}\label{sec:cost}

\begin{figure*}[t]
\centering
\includegraphics[width=\textwidth]{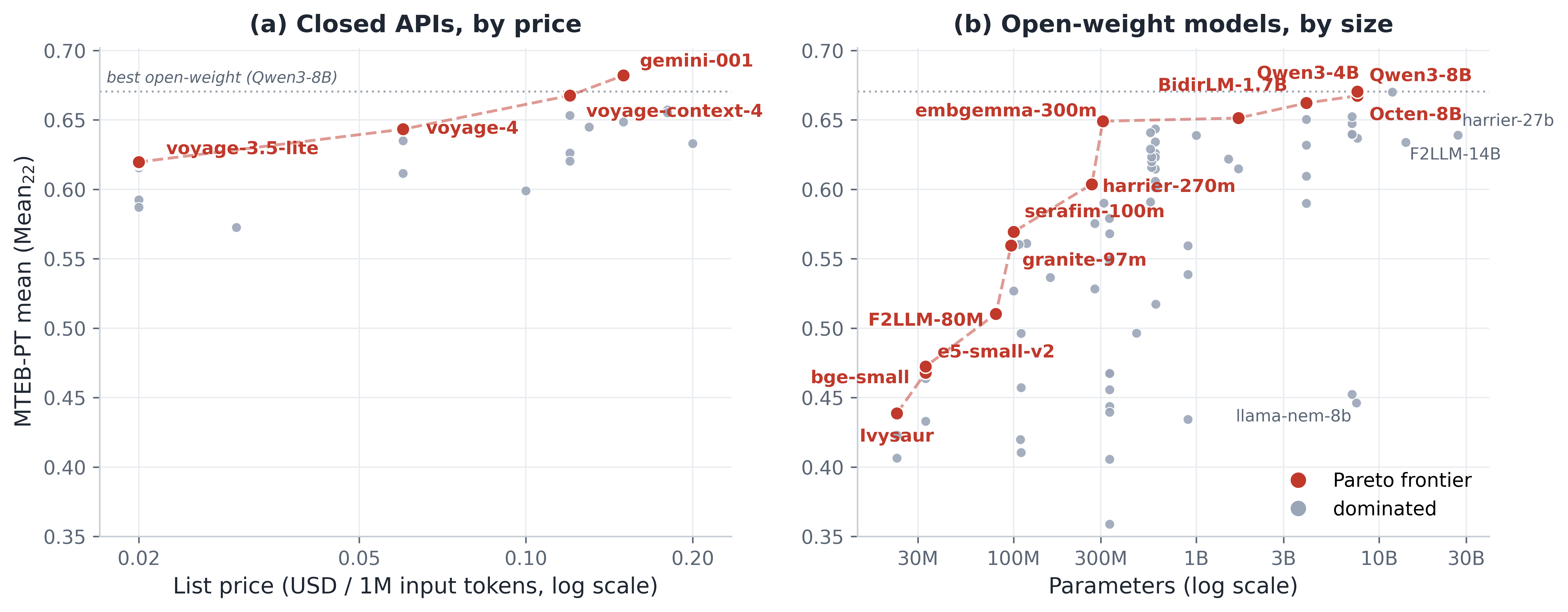}
\caption{Resource versus quality on MTEB-BR; both panels share the quality axis (22-task mean, $0.35$--$0.70$). \textbf{(a)} the 20 closed API endpoints against list price (USD per 1M input tokens, log scale); \textbf{(b)} the open-weight models against parameter count (log scale). Red points and the dashed line mark the Pareto frontier of each panel; gray points are dominated; the dotted horizontal line marks the best open-weight model (Qwen3-Embedding-8B), so its quality can be read directly against the priced endpoints in panel (a). Positions within a panel's frontier fall inside the converged frontier tier of \S\ref{sec:results} and are not statistically resolved from one another, so the frontier characterizes the quality--resource trade-off rather than a strict ordering of the top models. Closed-API list prices are provider snapshots as of July 2026.}
\label{fig:cost_pareto}
\end{figure*}

For deployment, the headline finding is that frontier quality does not require paying a per-token price: a free, self-hostable open-weight model, Qwen3-Embedding-8B (Apache-2.0, $0.670$ on the 22-task mean), reaches the converged frontier tier (\S\ref{sec:results}) alongside the top commercial APIs, at no per-token cost. The two classes are priced differently: a closed API charges per token, an open-weight model in the compute and memory its parameters demand. Both proxy the cost of producing an embedding, so we place each class on its own resource axis against the shared quality axis (Figure~\ref{fig:cost_pareto}): the 20 closed endpoints by list price (USD per 1M input tokens), the open-weight models by parameter count.

Among the 20 priced endpoints (Figure~\ref{fig:cost_pareto}a), four are Pareto-optimal: voyage-3.5-lite (\$0.02/1M, $0.620$), voyage-4 (\$0.06, $0.643$), voyage-context-4 (\$0.12, $0.668$), and gemini-embedding-001 (\$0.15, $0.682$); the other sixteen are dominated, and three of the four frontier points belong to Voyage. The frontier is shallow. A $7.5\times$ increase from the cheapest to the top-scoring frontier endpoint (\$0.02 to \$0.15) buys only $0.062$ in the 22-task mean, about three times the converged-frontier spread of \S\ref{sec:results}, and pricing rather than scale shapes it: text-embedding-3-large (\$0.13, $0.645$) is dominated outright by voyage-context-4, which is both cheaper and better. These positions hold under the 512-token evaluation cap (\S\ref{sec:protocol}), which leaves voyage-context-4's defining long-context capability inactive, so its frontier rank reflects short-input quality rather than the contextualized long-document embedding it is priced for. List prices are vendor-controlled and change without notice; the finding here is the structure of the frontier, its shallowness and the identity of the dominated endpoints, not the absolute dollar values, which we report as a dated snapshot.

The open-weight panel (Figure~\ref{fig:cost_pareto}b) shows the effect of model size. Quality rises steeply with scale at the small end and then flattens: the frontier climbs from sub-100M encoders to a knee at embeddinggemma-300m ($0.649$), beyond which two further orders of magnitude in parameters add little, and the largest open models (harrier-27b, F2LLM-14B, llama-embed-nemotron-8b) fall below the frontier rather than extending it. Scale buys retrieval quality on average (\S\ref{sec:discussion}) but is neither necessary nor sufficient at the top: a 0.3B model reaches near-frontier quality, and the frontier plateaus from there through the multi-billion range that Qwen3-Embedding-8B tops.

Read together, the two panels locate the models worth comparing and reduce the choice to operational axes. Within each frontier the scores fall inside the converged band of \S\ref{sec:results}, so the frontier orders the resource axes, not the models. The benchmark does not resolve the models' pairwise order there, and the decision then turns on latency, context length, region, and license. License in particular constrains the open option: several otherwise-competitive open models (Linq-Embed-Mistral, SFR-Embedding, Jina-v5) ship under non-commercial terms that bar production use, whereas Qwen3-Embedding is Apache-2.0. Where self-hosting is acceptable, an Apache-2.0 model at the frontier removes per-token cost entirely; where it is not, the priced frontier is shallow enough that the cheapest Pareto endpoint sits close to the best.

\section{Discussion, Limitations, and Ethics}\label{sec:discussion}

\paragraph{Discussion.} MTEB-BR discriminates the model field sharply. It resolves about a dozen distinguishable tiers and reports each with calibrated uncertainty rather than a bare point estimate. It is precise enough to establish that the six frontier models are statistically indistinguishable: a point-estimate ranking would invent an order among models that the benchmark places in a single tier. For the practitioner this makes the decision more precise, not less. Across the field the benchmark settles the ranking; within the frontier tier it identifies the models worth comparing on operational axes it does not measure, such as cost and license. And it answers that comparison favorably for open models, since a free, self-hostable open-weight model reaches the frontier tier at no per-token cost.

A transferable component of this work is the statistical layer. It lets the benchmark report a genuine tie as a tie and a real gap as a gap, and as embedding models cluster near the top in every language, uncertainty-aware evaluation becomes necessary rather than optional. We offer the layer, bootstrap intervals, paired significance, tier and probability-of-best estimates, IRT discrimination, and a Borda check, as a template for any language-specific MTEB extension, and we will contribute it upstream. Its discrimination estimates also point toward cheaper evaluation: the instance-level analysis demonstrates it directly, recovering the ranking from the most informative instances within a task, and the task-level discriminations suggest the same economy in choosing tasks, though we screen for discrimination rather than separately demonstrate task-subset recovery.

Finally, the results chart a path to better Portuguese embeddings. The apparent deficit of Portuguese-specific encoders on retrieval is a training-objective effect, not an effect of language: retrieval-tuned encoders on the same language match the multilingual models, while masked-language-model and legal-domain encoders trail. The productive route to stronger native retrieval is to fine-tune capable multilingual or decoder-LLM backbones on native Portuguese retrieval data; the mirror-image strength of those same encoders on classification shows that model choice should follow the task.

A further result concerns specialization. The apparent retrieval deficit of Portuguese-specific encoders (\S\ref{sec:results}, $\Delta = -0.186$, $p = 0.003$) is confounded by training objective: once the comparison is restricted to the general-purpose, retrieval-tuned Serafim encoders the gap is no longer detectable ($\Delta \approx 0$, $p = 0.99$). With only three IR-tuned Serafim variants this is an underpowered null, not a demonstration of equivalence; moreover the group $t$-test treats models as independent, but the panel is a convenience sample with several correlated families (multiple Serafim and Legal-BERTimbau variants share checkpoints and training data), so its effective sample size is smaller than the model count and the reported $p$-values are optimistic. We therefore read this as relocating the apparent language gap onto training objective and domain, not as proving that language-specific pretraining is irrelevant. In our panel, parameter count correlates with retrieval quality (Spearman $\rho = +0.63$ over the open models), consistent with the scale and breadth of contrastive training, rather than the language label, governing retrieval here; scale is nonetheless confounded with architecture, since the larger models are decoder-LLM embedders, and is neither necessary nor sufficient, with a 0.3B model outranking several encoders above 5B parameters.

\paragraph{Limitations.}
\begin{itemize}[leftmargin=*, itemsep=2pt, topsep=2pt]
\item \emph{Translation exclusion:} our native-source filter drops mMARCO-PT and mkqa-PT, the two largest machine-translated Portuguese retrieval corpora, which preserves signal but narrows retrieval-corpus diversity; we argue the trade-off is worth it (\S\ref{sec:benchmark}) while acknowledging it is contestable.
\item \emph{Document-length bias:} median document lengths are short (186 words for Quati, 116 for FaQuAD-IR, 43 for both JurisTCU and MedPTRetrieval), and only BR-TaxQA-R, at a 752-word median, exercises long-document retrieval; because we cap every input at 512 tokens (Appendix~\ref{app:repro}), even its documents are truncated, so models tuned for long contexts cannot show that advantage and are not stress-tested. A controlled check bounds the cost: on BR-TaxQA-R, lifting one model's (bge-m3) cap from 512 to 2048 tokens raised its nDCG@10 by nearly a quarter, so the uniform cap has a measurable long-document penalty that we accept in exchange for cross-model comparability. Since retrieval is the category that most separates models (\S\ref{sec:irt}) and a frontier endpoint, voyage-context-4, is a long-context model, the cap conditions both the retrieval ranking and the cost-Pareto, not only an unmeasured ceiling.
\item \emph{Small retrieval corpora:} some retrieval tasks index small candidate pools (FaQuAD-IR over 244 paragraphs, MedPTRetrieval over 500), so their nDCG@10 measures ranking within a small set rather than open-domain search and is easier than large-corpus retrieval; that said, the IRT discrimination of \S\ref{sec:irt} places both among the more separating tasks, so the small corpus does not flatten them.
\item \emph{Register skew:} the suite leans toward institutional and legal Portuguese. The JurisTCU corpus recurs three times (retrieval, reranking, clustering), and together with BR-TaxQA-R, PortuLexRRIP, and FaqBacen the legal and regulatory register accounts for six of the twenty-two formulations. Conversational, dialogue, and product-review text are not represented, so a score here speaks less directly to those deployment settings.
\item \emph{Closed-API drift:} closed APIs evolve silently; our scores are pinned to vendor identifiers and an evaluation window (closed-API queries and list prices are from June--July 2026), and should be re-checked at deployment.
\item \emph{Pretraining contamination:} several corpora are public web text (the Wikipedia-category, SciELO, and Stack Overflow clustering sets) and are likely present in the pretraining data of many evaluated models, so scores there may partly reflect memorized content rather than genuine embedding quality; the instance-level misfit audit (\S\ref{sec:irt}) targets mislabeling, not memorization, so we cannot rule this out for the web-sourced tasks.
\item \emph{Bootstrap resolution:} the clustering tasks have only 10 $k$-means experiments per model, limiting per-task bootstrap power, though the headline findings are robust to it.
\item \emph{Small evaluation sets:} the retrieval and reranking tasks built on Quati rest on 50 queries, and several classification probes on a few hundred test instances; the per-instance bootstrap CIs reflect this, and the smallest tasks carry correspondingly wide intervals.
\item \emph{Descriptive group tests:} the model panel is a convenience sample of available encoders rather than a random draw, so group-level comparisons such as Portuguese-specific versus multilingual (\S\ref{sec:results}) describe this panel rather than a population of models.
\end{itemize}

\paragraph{Ethics.} The content-moderation tasks (HateBR, ToxSynPT) contain offensive text; we redistribute it as the dataset authors do, without further filtering, and recommend human-in-the-loop screening before any downstream deployment. The social-media source (HateBR, drawn from Instagram) may carry personal identifiers present in the original posts; we redistribute it only under the source dataset's license and terms, add no new personal data, and recommend the same care before deployment. No native-speaker labor was uniquely employed beyond the original dataset authors', credited to the original sources in \S\ref{sec:benchmark} and Table~\ref{tab:per_task}. All code (Apache-2.0) and results (CC-BY 4.0) are released openly.

\section{Conclusion and Future Work}\label{sec:conclusion}

We have introduced MTEB-BR, a native Brazilian-Portuguese text-embedding benchmark of 22 tasks across seven MTEB categories and six domains, curated from existing Portuguese resources and excluding machine-translated corpora by construction. We evaluate 93 models on it, 73 open-weight (from 23M to 27B parameters) and 20 closed commercial APIs. Its distinguishing element is a statistical layer that reports the leaderboard as a measurement with explicit uncertainty rather than a bare ranking: per-task bootstrap confidence intervals, paired-bootstrap significance for every headline comparison, Item Response Theory discrimination at the task and instance levels, and a Borda-count robustness check.

Several findings bear directly on the model-selection decision. The 22-task mean spans 43 points, from $0.25$ to $0.68$, and the benchmark cleanly orders $78.7\%$ of model pairs into about a dozen tiers, placing the leader above 87 of 92 models. The six leaders fall within $0.020$ and converge into one frontier tier that gemini leads but the benchmark does not resolve, so within that tier the choice turns on cost and license rather than score. A self-hostable open-weight model sits inside that leading cluster at no per-token cost, and a 308M model ranks 13th of 93, within $0.033$ of the leader, placing near-frontier quality at edge-deployable size. The apparent deficit of Portuguese-specific encoders on retrieval tracks training objective and domain rather than language. And a model's rank on the global multilingual leaderboard predicts its Portuguese rank only moderately ($\rho = 0.75$ over 55 shared models, one of them 3rd there but 49th here), so a native benchmark measures something the multilingual boards do not.

These results come with limits: the panel is a convenience sample evaluated at a single point in time, the benchmark covers Brazilian Portuguese rather than the wider lusophone world, and it establishes that native and multilingual boards measure partly different constructs without adjudicating which is more deployment-relevant. The Item Response Theory layer also points beyond this benchmark, supporting adaptive evaluation that ranks models from fewer, higher-discrimination tasks and an overfitting diagnostic that tracks discrimination drift on near-duplicate data, both applicable to MTEB at large; natural next steps include a European-Portuguese dialect axis and contributing the statistical layer upstream to the \texttt{mteb} library as a reusable component.

For Brazilian Portuguese, a language of over 200 million speakers that until now had no consolidated native embedding benchmark, MTEB-BR lets practitioners choose models on native evidence rather than on translated or English-centric proxies, and gives researchers an extensible task suite and a reusable statistical template. All code (Apache-2.0), the full results dataset (CC-BY 4.0), and an interactive leaderboard are released openly under permanent DOIs, and we intend to keep the leaderboard current as new models appear.

\section*{Data and Code Availability}
The evaluation code is released under Apache-2.0 and archived at \url{https://doi.org/10.5281/zenodo.21087216}. The full results dataset, comprising per-model per-task scores and per-query metrics, is released under CC-BY 4.0 and archived at \url{https://doi.org/10.57967/hf/9491}. An interactive leaderboard is available at \url{https://huggingface.co/spaces/mteb-br/leaderboard}, and the task suite is distributed as an installable extension of the \texttt{mteb} library.

\section*{Acknowledgment}
We gratefully acknowledge Verda for the GPU compute credits that supported this work.

\appendices
\section{Reproducibility Details}\label{app:repro}
All evaluations are inference-only. Models are encoded on a single NVIDIA A100-80GB (Verda); the 20 closed APIs are called over HTTP and use no local GPU. We fix \texttt{max\_seq\_length} to 512 tokens across all models and tasks, encode in the model's native precision, and apply each model's documented pooling and padding configuration (mean, CLS, or last-token pooling, and left or right padding, per the model card; we verified the canonical configuration for every decoder-LLM embedder, where the wrong padding side silently degrades scores). Instruction-tuned models receive their default task-type prompt as shipped in the pinned \texttt{mteb} version; we add no custom instructions, so the prompt configuration is fixed by the model card and the \texttt{mteb} release rather than chosen by us. Every dataset is pinned to a HuggingFace revision SHA in its task wrapper. Classification uses a logistic-regression probe over an 80/20 stratified split with 10 resamples (the \texttt{mteb} default); clustering uses mini-batch $k$-means with \texttt{n\_init}\,$=10$ over 10 random seeds (\texttt{random\_state}\,$=0\ldots9$). Per-instance bootstrap confidence intervals use $10{,}000$ resamples; paired-bootstrap $p$-values use the per-task or per-instance difference vectors. We evaluated all models on \texttt{mteb} 2.12 (pair-classification accuracy computed per our released code); a three-seed check on HateBR (two models) returned identical scores across all three seeds, confirming the classification evaluator is deterministic at fixed batch size on GPU. Closed-API scores were collected against the vendor identifiers in Table~\ref{tab:model_catalog} during the evaluation window and should be re-checked at deployment, since endpoints evolve silently.

\section{Full Score Matrix}\label{app:matrix}
Table~\ref{tab:full_matrix} gives the per-task scores for the top 30 of 93 models by 22-task mean; the complete 93-model matrix is released with the results and on the interactive leaderboard.
\begin{table*}[t]
\centering
\caption{Per-task score matrix for the top 30 of 93 models by 22-task mean (each task's primary metric); task codes follow Table~\ref{tab:per_task}. The complete 93-model matrix is on the interactive leaderboard (see Data and Code Availability). The final row is a random-embedding baseline, whose per-task floor varies by category, so the $0.18$ aggregate is a mean over heterogeneous chance levels, not a single floor.}
\label{tab:full_matrix}
\scriptsize
\setlength{\tabcolsep}{1.2pt}
\resizebox{\textwidth}{!}{%
\begin{tabular}{lrrrrrrrrrrrrrrrrrrrrrrr}
\toprule
Model & HatBR & Fck & ToxPT & AsRTE & InfBR & AsSTS & As2ST & MedCl & WikCl & JurCl & SciCl & StkCl & MedRt & FaQIR & Quati & FaqBc & JurRt & BRTax & QuRrk & JuRrk & PtLex & BrEmo & \textbf{Mean} \\
\midrule
gemini-embedding-001 & 0.876 & 0.564 & 0.890 & 0.829 & 0.839 & 0.818 & 0.823 & 0.792 & 0.678 & 0.321 & 0.806 & 0.588 & 0.877 & 0.856 & 0.655 & 0.753 & 0.610 & 0.366 & 0.655 & 0.555 & 0.450 & 0.400 & \textbf{0.682} \\
Qwen3-Embedding-8B & 0.838 & 0.541 & 0.874 & 0.866 & 0.908 & 0.787 & 0.827 & 0.720 & 0.799 & 0.368 & 0.778 & 0.577 & 0.825 & 0.838 & 0.641 & 0.705 & 0.621 & 0.424 & 0.630 & 0.560 & 0.345 & 0.277 & \textbf{0.670} \\
KaLM-Embedding-Gemma3-12B-2511 & 0.856 & 0.572 & 0.901 & 0.886 & 0.899 & 0.806 & 0.818 & 0.733 & 0.772 & 0.370 & 0.826 & 0.524 & 0.874 & 0.831 & 0.560 & 0.730 & 0.583 & 0.364 & 0.622 & 0.537 & 0.397 & 0.283 & \textbf{0.670} \\
voyage-context-4 & 0.861 & 0.565 & 0.901 & 0.841 & 0.811 & 0.786 & 0.802 & 0.660 & 0.605 & 0.429 & 0.660 & 0.449 & 0.892 & 0.874 & 0.690 & 0.806 & 0.634 & 0.386 & 0.686 & 0.577 & 0.473 & 0.298 & \textbf{0.668} \\
Octen-Embedding-8B & 0.835 & 0.532 & 0.875 & 0.864 & 0.910 & 0.786 & 0.823 & 0.723 & 0.792 & 0.375 & 0.778 & 0.574 & 0.849 & 0.823 & 0.644 & 0.717 & 0.601 & 0.373 & 0.634 & 0.543 & 0.356 & 0.277 & \textbf{0.667} \\
Qwen3-Embedding-4B & 0.819 & 0.505 & 0.830 & 0.866 & 0.894 & 0.793 & 0.821 & 0.886 & 0.726 & 0.411 & 0.741 & 0.626 & 0.773 & 0.810 & 0.610 & 0.655 & 0.616 & 0.420 & 0.598 & 0.554 & 0.328 & 0.286 & \textbf{0.662} \\
voyage-context-3 & 0.848 & 0.571 & 0.888 & 0.839 & 0.770 & 0.777 & 0.776 & 0.645 & 0.655 & 0.410 & 0.648 & 0.486 & 0.891 & 0.853 & 0.671 & 0.796 & 0.624 & 0.330 & 0.671 & 0.575 & 0.444 & 0.287 & \textbf{0.657} \\
voyage-3-large & 0.848 & 0.583 & 0.887 & 0.804 & 0.776 & 0.750 & 0.781 & 0.642 & 0.635 & 0.425 & 0.669 & 0.434 & 0.881 & 0.863 & 0.668 & 0.799 & 0.616 & 0.384 & 0.670 & 0.568 & 0.461 & 0.271 & \textbf{0.655} \\
voyage-4-large & 0.842 & 0.540 & 0.893 & 0.800 & 0.789 & 0.762 & 0.768 & 0.798 & 0.538 & 0.456 & 0.661 & 0.542 & 0.894 & 0.858 & 0.684 & 0.734 & 0.547 & 0.364 & 0.666 & 0.515 & 0.434 & 0.286 & \textbf{0.653} \\
SFR-Embedding-Mistral & 0.816 & 0.447 & 0.896 & 0.862 & 0.845 & 0.804 & 0.799 & 0.807 & 0.709 & 0.307 & 0.721 & 0.653 & 0.817 & 0.830 & 0.626 & 0.698 & 0.541 & 0.350 & 0.625 & 0.531 & 0.357 & 0.309 & \textbf{0.652} \\
BidirLM-1.7B-Embedding & 0.823 & 0.502 & 0.898 & 0.855 & 0.859 & 0.766 & 0.800 & 0.857 & 0.740 & 0.318 & 0.781 & 0.629 & 0.831 & 0.758 & 0.569 & 0.673 & 0.548 & 0.266 & 0.586 & 0.526 & 0.446 & 0.298 & \textbf{0.651} \\
BOOM\_4B\_v1 & 0.852 & 0.556 & 0.853 & 0.856 & 0.769 & 0.754 & 0.742 & 0.776 & 0.733 & 0.318 & 0.766 & 0.546 & 0.813 & 0.862 & 0.607 & 0.724 & 0.641 & 0.327 & 0.598 & 0.560 & 0.378 & 0.277 & \textbf{0.650} \\
embeddinggemma-300m & 0.830 & 0.569 & 0.859 & 0.876 & 0.873 & 0.789 & 0.799 & 0.719 & 0.686 & 0.294 & 0.700 & 0.545 & 0.777 & 0.846 & 0.607 & 0.695 & 0.621 & 0.375 & 0.569 & 0.502 & 0.422 & 0.324 & \textbf{0.649} \\
codestral-embed & 0.834 & 0.570 & 0.879 & 0.841 & 0.864 & 0.806 & 0.811 & 0.659 & 0.619 & 0.276 & 0.675 & 0.449 & 0.887 & 0.812 & 0.594 & 0.826 & 0.637 & 0.327 & 0.606 & 0.579 & 0.431 & 0.285 & \textbf{0.649} \\
Linq-Embed-Mistral & 0.827 & 0.454 & 0.908 & 0.856 & 0.851 & 0.803 & 0.798 & 0.701 & 0.757 & 0.233 & 0.682 & 0.567 & 0.815 & 0.840 & 0.604 & 0.742 & 0.623 & 0.272 & 0.618 & 0.581 & 0.417 & 0.292 & \textbf{0.647} \\
text-embedding-3-large & 0.863 & 0.552 & 0.898 & 0.827 & 0.756 & 0.792 & 0.761 & 0.722 & 0.614 & 0.323 & 0.700 & 0.509 & 0.846 & 0.789 & 0.636 & 0.746 & 0.609 & 0.269 & 0.639 & 0.558 & 0.480 & 0.299 & \textbf{0.645} \\
jina-embeddings-v5-text-small & 0.797 & 0.435 & 0.774 & 0.858 & 0.851 & 0.777 & 0.816 & 0.799 & 0.758 & 0.340 & 0.765 & 0.617 & 0.780 & 0.820 & 0.594 & 0.687 & 0.602 & 0.360 & 0.624 & 0.558 & 0.287 & 0.260 & \textbf{0.643} \\
voyage-4 & 0.839 & 0.534 & 0.890 & 0.811 & 0.784 & 0.738 & 0.800 & 0.749 & 0.550 & 0.456 & 0.666 & 0.491 & 0.880 & 0.845 & 0.628 & 0.737 & 0.544 & 0.370 & 0.636 & 0.511 & 0.407 & 0.289 & \textbf{0.643} \\
multilingual-e5-large-instruct & 0.826 & 0.557 & 0.894 & 0.877 & 0.827 & 0.808 & 0.806 & 0.723 & 0.788 & 0.367 & 0.721 & 0.506 & 0.769 & 0.812 & 0.561 & 0.692 & 0.579 & 0.182 & 0.593 & 0.559 & 0.321 & 0.330 & \textbf{0.641} \\
SFR-Embedding-2\_R & 0.815 & 0.446 & 0.892 & 0.851 & 0.733 & 0.776 & 0.745 & 0.766 & 0.762 & 0.238 & 0.719 & 0.636 & 0.802 & 0.739 & 0.565 & 0.694 & 0.612 & 0.394 & 0.563 & 0.575 & 0.425 & 0.326 & \textbf{0.640} \\
gte-Qwen2-7B-instruct & 0.841 & 0.563 & 0.879 & 0.846 & 0.829 & 0.782 & 0.794 & 0.729 & 0.766 & 0.308 & 0.760 & 0.457 & 0.819 & 0.779 & 0.581 & 0.694 & 0.541 & 0.300 & 0.580 & 0.516 & 0.387 & 0.310 & \textbf{0.639} \\
harrier-oss-v1-27b & 0.877 & 0.605 & 0.886 & 0.794 & 0.797 & 0.786 & 0.835 & 0.649 & 0.619 & 0.447 & 0.701 & 0.487 & 0.881 & 0.732 & 0.563 & 0.708 & 0.544 & 0.331 & 0.574 & 0.512 & 0.474 & 0.255 & \textbf{0.639} \\
BidirLM-1B-Embedding & 0.822 & 0.493 & 0.899 & 0.864 & 0.854 & 0.783 & 0.803 & 0.716 & 0.759 & 0.347 & 0.724 & 0.478 & 0.812 & 0.772 & 0.555 & 0.674 & 0.579 & 0.242 & 0.579 & 0.546 & 0.437 & 0.319 & \textbf{0.639} \\
F2LLM-v2-8B & 0.840 & 0.585 & 0.840 & 0.785 & 0.800 & 0.774 & 0.785 & 0.614 & 0.523 & 0.410 & 0.587 & 0.484 & 0.871 & 0.765 & 0.649 & 0.823 & 0.602 & 0.341 & 0.627 & 0.565 & 0.432 & 0.308 & \textbf{0.637} \\
voyage-3.5 & 0.828 & 0.534 & 0.867 & 0.804 & 0.736 & 0.730 & 0.749 & 0.663 & 0.667 & 0.374 & 0.720 & 0.477 & 0.833 & 0.847 & 0.615 & 0.751 & 0.591 & 0.343 & 0.633 & 0.558 & 0.398 & 0.251 & \textbf{0.635} \\
harrier-oss-v1-0.6b & 0.801 & 0.547 & 0.853 & 0.859 & 0.848 & 0.771 & 0.795 & 0.754 & 0.715 & 0.348 & 0.703 & 0.498 & 0.754 & 0.826 & 0.597 & 0.651 & 0.544 & 0.346 & 0.613 & 0.525 & 0.328 & 0.274 & \textbf{0.634} \\
F2LLM-v2-14B & 0.861 & 0.579 & 0.837 & 0.787 & 0.806 & 0.778 & 0.790 & 0.602 & 0.495 & 0.414 & 0.569 & 0.478 & 0.886 & 0.761 & 0.643 & 0.819 & 0.591 & 0.305 & 0.635 & 0.564 & 0.445 & 0.301 & \textbf{0.634} \\
gemini-embedding-2 & 0.850 & 0.570 & 0.896 & 0.769 & 0.691 & 0.748 & 0.727 & 0.735 & 0.615 & 0.295 & 0.731 & 0.494 & 0.875 & 0.860 & 0.647 & 0.775 & 0.489 & 0.294 & 0.643 & 0.478 & 0.494 & 0.251 & \textbf{0.633} \\
F2LLM-v2-4B & 0.841 & 0.581 & 0.836 & 0.780 & 0.776 & 0.760 & 0.778 & 0.748 & 0.483 & 0.438 & 0.549 & 0.480 & 0.863 & 0.758 & 0.616 & 0.814 & 0.590 & 0.309 & 0.614 & 0.561 & 0.427 & 0.294 & \textbf{0.632} \\
PwC-Embedding\_expr & 0.826 & 0.556 & 0.880 & 0.881 & 0.852 & 0.805 & 0.803 & 0.677 & 0.714 & 0.402 & 0.631 & 0.574 & 0.768 & 0.796 & 0.550 & 0.670 & 0.525 & 0.204 & 0.580 & 0.504 & 0.342 & 0.298 & \textbf{0.629} \\
\midrule
\textit{Random baseline} & 0.502 & 0.322 & 0.495 & 0.233 & 0.356 & 0.005 & -0.029 & 0.529 & 0.325 & 0.123 & 0.086 & 0.335 & 0.008 & 0.023 & 0.000 & 0.003 & 0.000 & 0.013 & 0.180 & 0.143 & 0.142 & 0.203 & \textit{0.182} \\
\bottomrule
\end{tabular}%
}
\end{table*}

\section{Construction of the Tasks Introduced Here}\label{app:construction}
Several headline tasks are constructed or reformulated by this work from raw Brazilian-Portuguese sources; we document their construction below.

\paragraph{WikiCatClusP2P} is constructed by us from the Brazilian-Portuguese Wikipedia (\texttt{wikimedia/wikipedia}, \texttt{20231101.pt} snapshot) via the MediaWiki \texttt{categorymembers} and \texttt{extracts} APIs. We traverse 15 broad subject categories (História, Geografia, Política, Esporte, Música, Cinema, Literatura, Religião, Ciência, Tecnologia, Animais, Plantas, Medicina, Filosofia, Astronomia) to depth 2, and take the first paragraph (between 80 and 500 characters) of each discovered article, labelled by the root category through which it was first reached. The released dataset (\texttt{mteb-br/wikipedia-categories}, CC-BY-SA-3.0 inherited from Wikipedia) has 2{,}873 article paragraphs across the 15 clusters. Label noise can arise when an article legitimately belongs to several categories; we keep the first-discovery label and report this as a known limitation.

\paragraph{MedPTRetrieval and MedPTClustering} reformulate the \texttt{AKCIT/MedPT} medical question--answer corpus (CC-BY-4.0) into MTEB task formats. For retrieval we draw a stratified sample of 500 question--answer pairs (\texttt{random\_state}\,$=42$, balanced across the 7 \texttt{question\_type} categories), using the 500 questions as queries and their 500 paired answers as the corpus with a $1{:}1$ gold mapping. For clustering we sample $\approx\!600$ questions across the 12 broad medical specialties (about 50 per specialty) and cluster by specialty.

\paragraph{SciELOClusteringP2P, StackoverflowPtClustering, and JurisTCUClusteringP2P} cluster native Brazilian-Portuguese texts by a coarse topic label drawn from each source's own taxonomy. SciELO clusters scientific abstracts from the SciELO Brazil open-access library into 8 broad research areas; StackoverflowPt clusters technical question titles from the Portuguese Stack Overflow (\texttt{pt.stackoverflow.com}) into 10 technology tags; JurisTCU clusters Federal Court of Accounts (TCU) jurisprudence excerpts into 10 legal areas. Each released dataset (\texttt{mteb-br/scielo-clustering}, \texttt{mteb-br/stackoverflow-clustering}, \texttt{mteb-br/juristcu-clustering}) carries the source label as the gold cluster.

\paragraph{FaqBacenRetrieval and FaQuADIR} are native retrieval tasks. FaqBacen pairs 373 citizen questions about Brazilian financial regulation with a pool of 1{,}673 unique answers from the Banco Central do Brasil public FAQ (\texttt{mteb-br/faq-bacen}); FaQuADIR reformulates FaQuAD as academic retrieval, mapping 900 questions about Brazilian higher education to 244 source paragraphs drawn from 18 official documents (\texttt{mteb-br/faquad-ir}). The two corpus-derived reranking tasks (\texttt{mteb-br/quati-reranking}, \texttt{mteb-br/juristcu-reranking}) reuse the Quati and JurisTCU corpora, restricting each query to its pool of human-judged candidate passages.

\section{Translated Retrieval and Train--Test Overlap}\label{app:mmarco}

We evaluate translated mMARCO-PT (\texttt{unicamp-dl/mmarco}, Google translation \cite{bonifacio-etal-2021-mmarco}) as a dense-retrieval task under the same nDCG@10 metric and encoding pipeline as the native retrieval tasks: 150 dev-small queries against a 30k-passage corpus of the relevant passages plus per-query BM25 top-200 hard negatives, over 73 open models spanning the ability range; the rank comparison below is over the 71 of these that also have native-retrieval scores. Because mMARCO-PT and the native tasks are encoded identically, each model's mMARCO-versus-native difference is a within-model comparison.

\begin{figure}[t]
\centering
\includegraphics[width=\columnwidth]{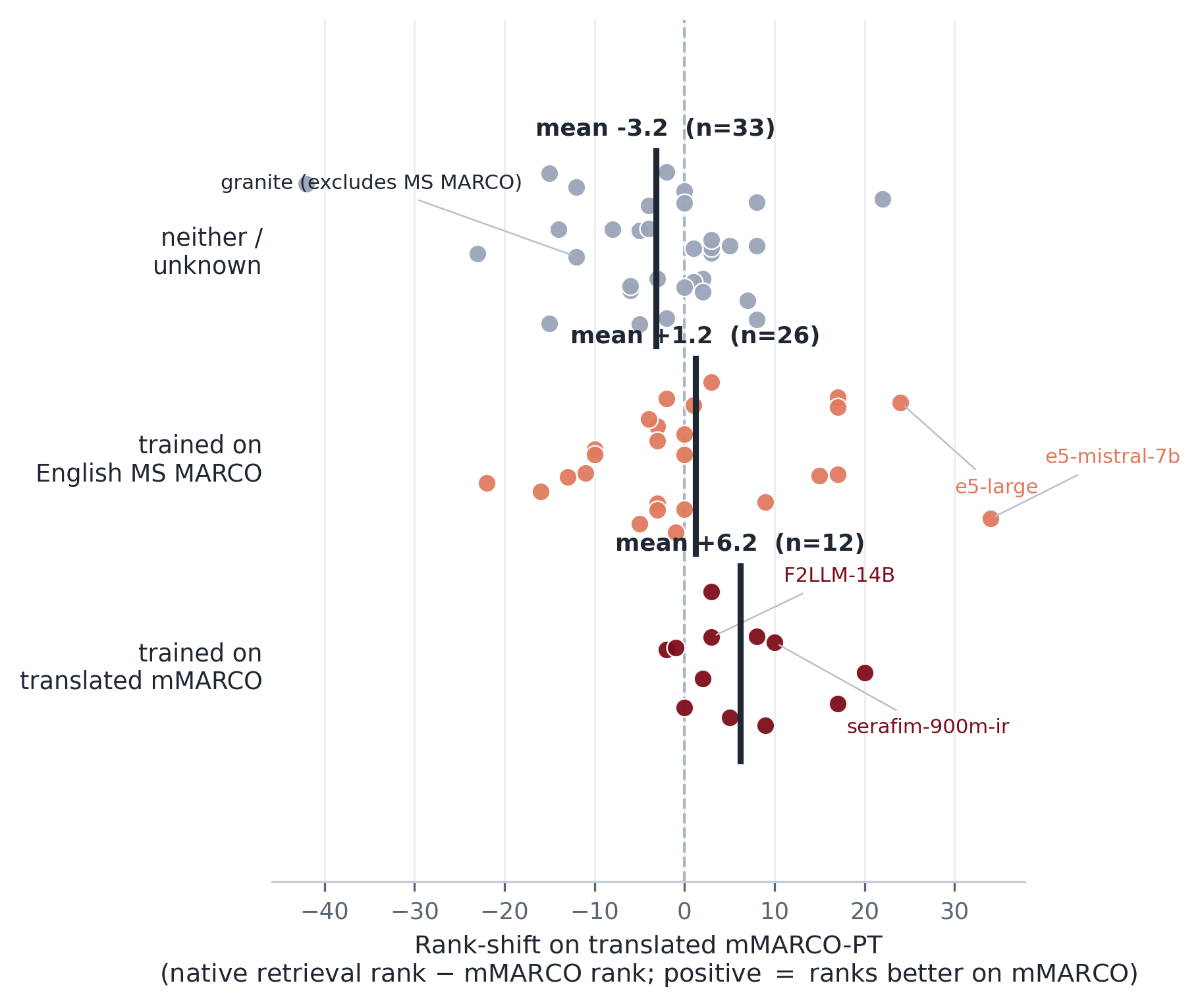}
\caption{Rank-shift on translated mMARCO-PT by training-data cohort (71 open models with native-retrieval scores). Positive means a model ranks better on mMARCO-PT than on the native retrieval tasks. Models trained on MS MARCO or its Portuguese translation over-perform; a model that excludes MS MARCO does not. Vertical bars are cohort means.}
\label{fig:mmarco}
\end{figure}

On this panel mMARCO-PT separates models, and its ranking correlates with native retrieval (Spearman $\rho = 0.84$). Within that agreement, scores are systematically higher for models trained on MS MARCO. Sorting models into three cohorts by declared training data yields a monotonic rank-shift (Figure~\ref{fig:mmarco}): models trained on the translated mMARCO, which includes the Portuguese split (F2LLM \cite{f2llm-2025}, and the Serafim-IR encoders, fine-tuned on the $\sim$40M-triple Brazilian-Portuguese portion \cite{rodrigues-etal-2025-serafim}), gain a mean of $+6.2$ ranks on mMARCO-PT relative to native retrieval; models trained on English MS MARCO gain $+1.2$; models trained on neither lose $-3.2$. F2LLM tops mMARCO-PT at every size while ranking mid-pack natively, and Serafim-IR over-performs while its non-IR sibling does not. A model whose training explicitly excludes MS MARCO \cite{ibm-2024-granite} ranks lower on mMARCO-PT than natively, the no-exposure control.

Cohorts are assigned from self-declared training data, so any undisclosed MS MARCO training falls into the ``neither'' group and narrows the observed gap; the effect is a lower bound. The web-sourced clustering tasks (\S\ref{sec:discussion}) carry a weaker, corpus-level exposure risk, since their text may appear in pretraining; but no model trains on their task signal, and native Portuguese is a small fraction of these models' predominantly English pretraining, so the translated-retrieval overlap is the more direct one.

\section{Inter-Task Ranking Agreement}\label{app:spearman}
\begin{figure}[t]
\centering
\includegraphics[width=\columnwidth]{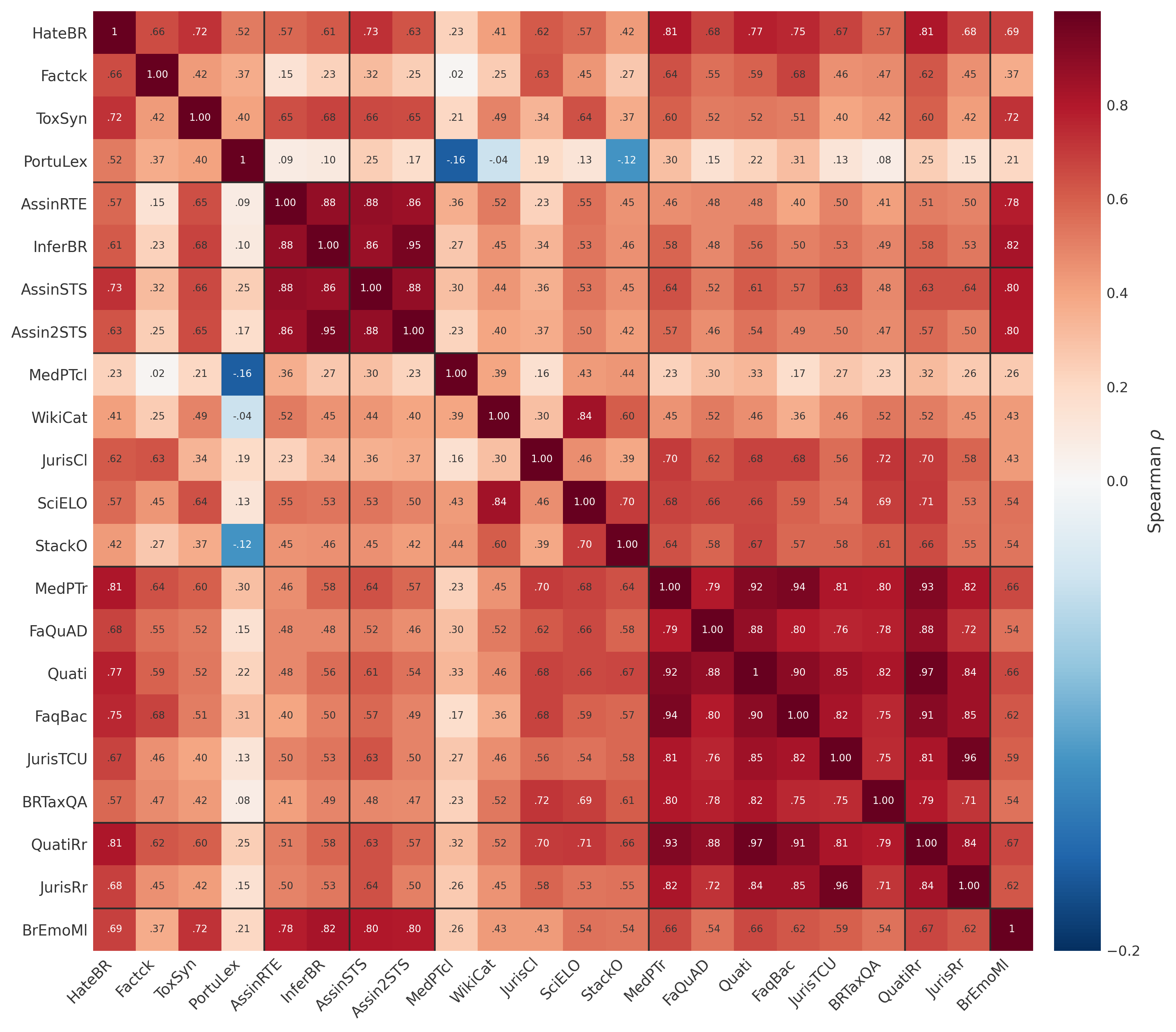}
\caption{\textbf{Model-ranking agreement.} Spearman rank correlation between the 22 headline tasks over the 93-model panel. Within-category agreement (mean $0.67$) exceeds across-category agreement ($0.51$); the clustering tasks are the least correlated with the rest of the suite.}
\label{fig:irt_spearman}
\end{figure}

Figure~\ref{fig:irt_spearman} gives the full Spearman rank-correlation matrix between the 22 headline tasks over the 93-model panel, the model-ranking-agreement diagnostic of \S\ref{sec:irt}. Retrieval and reranking tasks agree most (the $0.97$ Quati--QuatiReranking pair is the highest), while the clustering tasks are the least correlated with the rest of the suite.

\section{Statistical Significance at the Top}\label{app:significance}
Table~\ref{tab:top10_p} reports the paired-bootstrap two-sided $p$-values ($10{,}000$ resamples of the 22 headline tasks) for every pair among the ten leading models, the converged-tier detail behind \S\ref{sec:results}. Of the 45 comparisons among the top ten, 39 fail to reject an equal mean at $\alpha = 0.05$; of the six that do, all involve a model ranked seventh or lower, so no pair within the top six resolves. One survives a Holm--Bonferroni correction: voyage-context-4 over voyage-3-large ($p \approx 0.001$); the next-strongest, gemini over SFR-Embedding-Mistral ($p \approx 0.002$), sits just above the corrected threshold, and the four marginal rejections ($p \in [0.02, 0.05]$) are consistent with the $\approx 2$ false positives expected at $\alpha = 0.05$.
\begin{table}[t]
\centering
\caption{Paired-bootstrap two-sided $p$-values among the ten leading models (22-task mean in parentheses), from 10{,}000 resamples of the 22 headline tasks. Cells $\le0.05$ uncorrected are in bold. The top six are pairwise-unresolved (one converged tier); the appendix text gives the Holm--Bonferroni detail.}
\label{tab:top10_p}
\scriptsize
\setlength{\tabcolsep}{2.6pt}
\resizebox{\columnwidth}{!}{%
\begin{tabular}{lrrrrrrrrr}
\toprule
 & 1 & 2 & 3 & 4 & 5 & 6 & 7 & 8 & 9 \\
\midrule
2 Qwen3-Embedding-8B (0.670) & .30 &  &  &  &  &  &  &  &  \\
3 KaLM-Embedding-Gemma3-12B-2511 (0.670) & .25 & .96 &  &  &  &  &  &  &  \\
4 voyage-context-4 (0.668) & .28 & .89 & .91 &  &  &  &  &  &  \\
5 Octen-Embedding-8B (0.667) & .17 & .30 & .67 & .95 &  &  &  &  &  \\
6 Qwen3-Embedding-4B (0.662) & .15 & .40 & .56 & .78 & .60 &  &  &  &  \\
7 voyage-context-3 (0.657) & \textbf{.04} & .37 & .38 & \textbf{.03} & .47 & .80 &  &  &  \\
8 voyage-3-large (0.655) & \textbf{.05} & .33 & .33 & \textbf{.00} & .43 & .74 & .65 &  &  \\
9 voyage-4-large (0.653) & \textbf{.02} & .32 & .33 & .19 & .40 & .60 & .72 & .84 &  \\
10 SFR-Embedding-Mistral (0.652) & \textbf{.00} & .09 & .15 & .41 & .13 & .30 & .77 & .87 & .93 \\
\bottomrule
\end{tabular}
}
\end{table}

\section{Instance-Level IRT: Per-Task Results}\label{app:irt_instance}
Table~\ref{tab:irt_instance} reports the per-task numbers behind the instance-level analysis of \S\ref{sec:irt_instance}.

\begin{table}[htbp]
\centering
\caption{\textbf{Instance-level IRT, per-task results} (2-PL of \S\ref{sec:irt_instance}, four dichotomous tasks with released per-instance predictions). $\theta$--acc $\tau$: Kendall agreement of the top-$K{=}50$ highest-Fisher-information instances with the full-task ranking, against the mean of 400 random draws (\emph{Fisher / random}). \emph{Held-out}: cross-model validation (mean of 5 splits). \emph{Misfit}: fraction of items with negative item--ability correlation.}
\label{tab:irt_instance}
\small
\resizebox{\columnwidth}{!}{%
\begin{tabular}{lrrrrr}
\toprule
Task & $I\times M$ & $\theta$--acc $\tau$ & Fisher / random & held-out & Misfit \\
\midrule
ToxSynPT & 5208$\times$36 & 0.96 & 0.90 / 0.50 & 0.82 & 11\% \\
InferBR  & 1705$\times$35 & 0.86 & 0.77 / 0.65 & 0.72 & 26\% \\
HateBR   & 1400$\times$37 & 0.86 & 0.70 / 0.52 & 0.64 & 18\% \\
AssinRTE & 2000$\times$35 & 0.74 & 0.66 / 0.42 & 0.66 & 30\% \\
\bottomrule
\end{tabular}%
}
\end{table}

\clearpage

\makeatletter
\renewcommand{\@IEEEsectpunct}{.\ \,}
\def\footnotesize{\@setfontsize{\footnotesize}{8}{9.5pt}}
\makeatother

\raggedbottom

\bibliographystyle{IEEEtran}
\bibliography{references}

\end{document}